\documentclass[runningheads]{llncs}

\usepackage{eccv}

\usepackage{eccvabbrv}

\usepackage{graphicx}
\usepackage{booktabs}
\usepackage{comment}

\usepackage{caption}
\usepackage{multirow}
\usepackage{wrapfig}
\usepackage{subcaption}

\usepackage[accsupp]{axessibility}  

\newcommand*\samethanks[1][\value{footnote}]{\footnotemark[#1]}

\usepackage{hyperref}

\usepackage{orcidlink}

\begin{document}

\title{Bridging the Gap between Human Motion and Action Semantics via Kinematic Phrases} 

\titlerunning{Kinematic Phrases}

\author{Xinpeng Liu\inst{1}\orcidlink{0000-0002-7525-3243} \and
Yong-Lu Li\inst{1}\thanks{Corresponding authors.}\orcidlink{0000-0003-0478-0692} \and
Ailing Zeng\inst{2}\orcidlink{0000-0002-3783-0679} \and
Zizheng Zhou\inst{1} \and
Yang You\inst{3}\orcidlink{0000-0003-0125-0792} \and
Cewu Lu\inst{1}\samethanks\orcidlink{0000-0003-1533-8576}
}
\authorrunning{Liu et al.}
\institute{Shanghai Jiao Tong University \and
Tencent \and
Stanford University\\
\email{\{xinpengliu0907,ailingzengzzz\}@gmail.com, \{yonglu\_li,zhou\_zz,lucewu\}@sjtu.edu.cn, yangyou@stanford.edu} 
}
\maketitle

\begin{abstract}
Motion understanding aims to establish a reliable mapping between motion and action semantics, while it is a challenging many-to-many problem.
An abstract action semantic (i.e., \textit{walk forwards}) could be conveyed by perceptually diverse motions (walking with arms up or swinging). In contrast, a motion could carry different semantics w.r.t. its context and intention.
This makes an elegant mapping between them difficult.
Previous attempts adopted direct-mapping paradigms with limited reliability.
Also, current automatic metrics fail to provide reliable assessments of the consistency between motions and action semantics.
We identify the source of these problems as the \textbf{significant gap} between the two modalities.
To alleviate this gap, we propose Kinematic Phrases (KP) that take the objective kinematic facts of human motion with \textbf{proper abstraction}, \textbf{interpretability}, and \textbf{generality}. 
Based on KP, we can unify a motion knowledge base and build a motion understanding system.
Meanwhile, KP can be \textit{automatically} converted from motions to text descriptions with no subjective bias, inspiring Kinematic Prompt Generation (KPG) as a \textbf{novel white-box motion generation benchmark}. 
In extensive experiments, our approach shows superiority over other methods. 
Our project is available at \url{https://foruck.github.io/KP/}.

\end{abstract}

\section{Introduction}
Human motion understanding has a wide range of applications, including activity understanding~\cite{li2024isolated},  autonomous driving~\cite{paden2016survey}, robotics~\cite{koppula2013anticipating}, and automatic animation~\cite{van2010real}, making it increasingly attractive.
The core is to establish a mapping between the motion space and the action semantics space.
The motion space indicates a space of sequential 3D human representations, e.g., 3D pose or SMPL~\cite{smpl}/SMPL-X~\cite{smplx} parameter sequence, while the action semantic space can be represented as action categories or sentences described by natural language.

Recently, a growing focus has been on generative mapping from semantics to motion, including action category-based generation~\cite{actor} and text-based generation~\cite{temos,hml3d,posegpt,motiondiffuse,hmdm,mld,t2mgpt}.
Most of them typically build a mapping that links motion and semantics either directly or via motion latent, with understated concerns for intermediate motion-semantic structures.
However, these models suffer from inferior reliability.
They cannot guarantee they generated correct samples without human filtering.
Additionally, the existing evaluation of motion generation is problematic. 
Widely adopted FID and R-Precision rely on the latent space from a black-box pre-trained model, which might fail to out-of-distribution (OOD) and over-fitting cases.
There is a long-standing need for an evaluation method that can cheaply and reliably assess whether a generated motion is consistent with particular action semantics. 
We identify the essence of these as the significant gap between raw human motion and action semantics, which makes direct mapping hard to learn.

\begin{figure}[t]
    \hsize=\textwidth
    \centering
    \includegraphics[width=\textwidth]{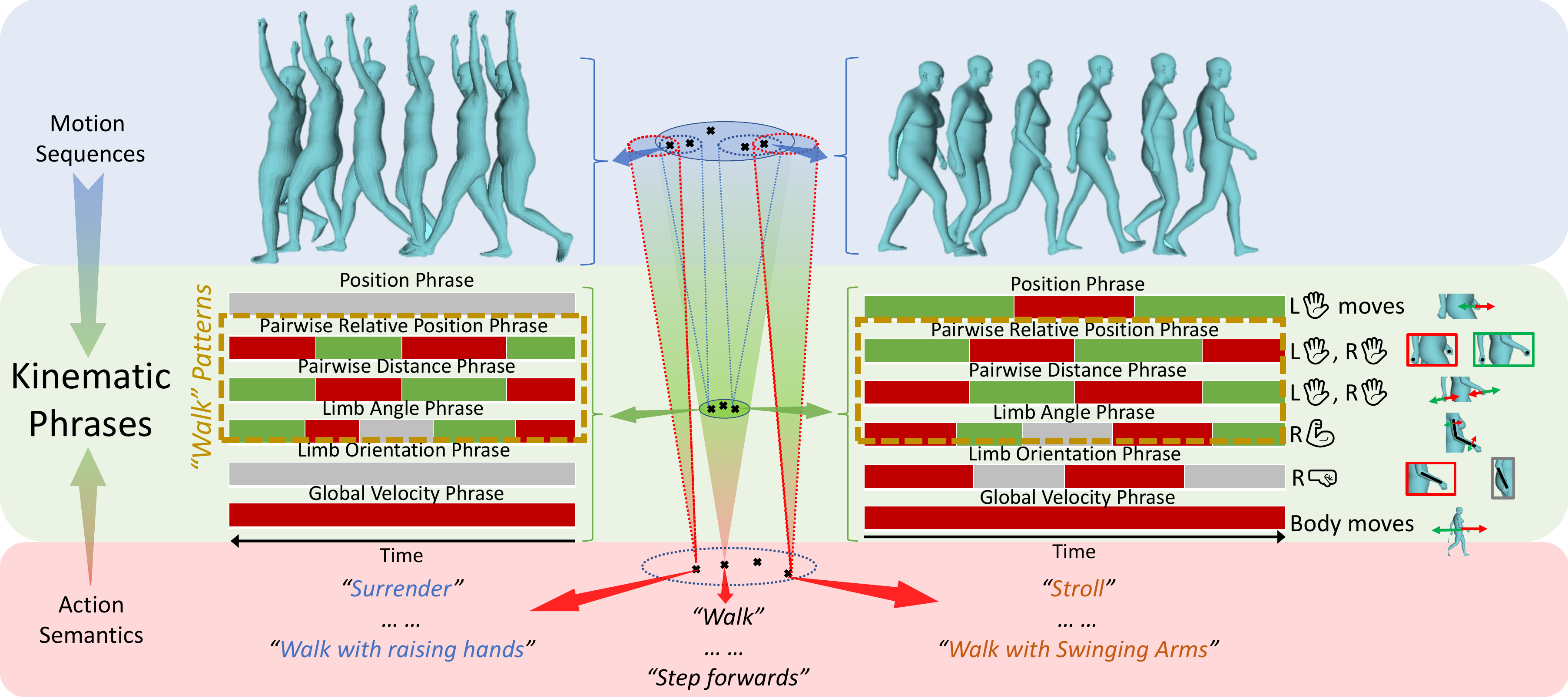}
    \caption{
    The huge gap between motion and action semantics results in the \textit{many-to-many} problem. 
    We propose Kinematic Phrases (KP) as an intermediate to bridge the gap.
    KPs objectively capture human kinematic cues.  
    It properly abstracts diverse motions with interpretability.
    As shown, the Phrases in the yellow box could capture key patterns of \textit{walk} for diverse motions.
    } 
    \label{fig:1}
\end{figure}

As in Fig.~\ref{fig:1}, an action semantics can correspond to diverse motions. 
For instance, a person could \textit{walk} in countless ways with diverse motions, either with arms up or swinging, while action semantics tend to abstract these away from a walking motion.
Additionally, they are robust against small perturbations, while motion is more specific and complex, with representations changing vastly when perturbed or mis-captured.
Moreover, a motion sequence could have diverse semantics w.r.t. contexts. 
Modeling this many-to-many mapping between motion and semantics is challenging. 

To bridge this gap between motion and action semantics, we propose Kinematic Phrases (KP), an interpretable intermediate representation. 
KP focuses on the objective kinematic facts, which are usually omitted by general action semantics, like \texttt{left-hand moving forwards then backward}.
KP is designed as qualitative categorical representations of these facts.
For objectivity and actuality, KP captures \textbf{sign changes} with minimal pre-defined standards. 
Inspired by previous studies on kinematic human motion representation~\cite{von1975laban,bartlett1997introduction}, KP is proposed 
\begin{wrapfigure}{r}{0.5\textwidth}
    \centering
    \includegraphics[width=0.48\textwidth]{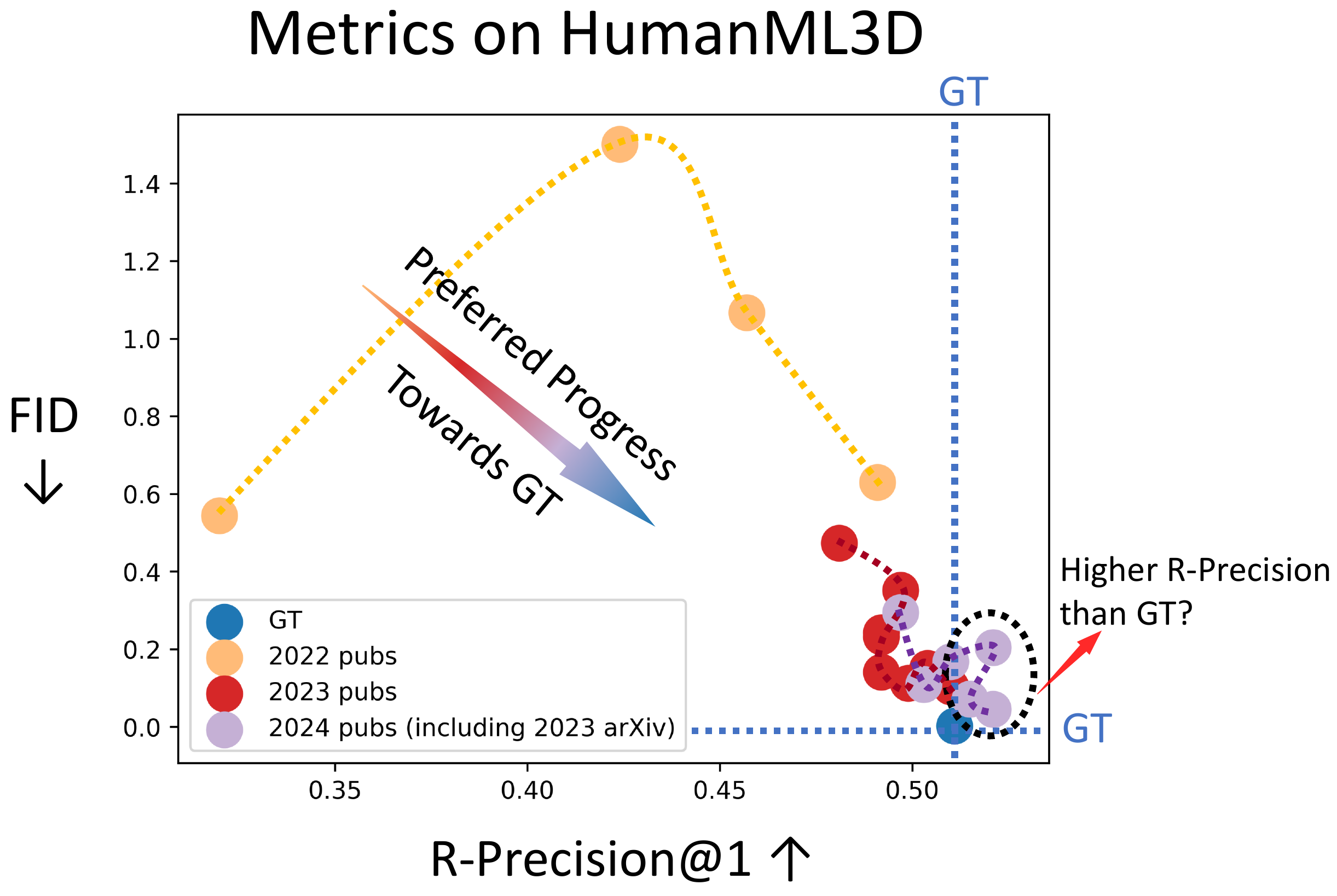}
    \caption{HumanML3D~\cite{hml3d} metric values are approaching even surpassing GT-level while being increasingly indistinguishable. The higher-than-GT R-Precision could be insufficient as a semantic consistency indicator.}
    \label{fig:bench}
\end{wrapfigure}
as six types shown in Fig.~\ref{fig:1}, covering \textbf{joint positions}, \textbf{joint pair positions} and \textbf{distances}, \textbf{limb angles} and \textbf{directions}, and \textbf{global velocity}. 
Note that, although KP can be described by natural language, a major difference is that KP is strictly dedicated to objective kinematic facts instead of coarse actions such as \textit{surrender} or fine-grained actions like \textit{raise both hands}.

We highlight three advantages of KP. 
First, KP offers \textbf{proper abstraction}, which disentangles motion perturbations and semantics changes, easing the learning process.
Even though the motion differs significantly, KP manages to capture \textit{walk} patterns easily.
Second, KP is \textbf{interpretable}, as it can be viewed as instructions on executing the action, making it easily understandable to humans. 
Finally, KP is \textbf{general}, as it can be automatically extracted from different modalities of human motion, including skeleton and SMPL parameters. The conversion from KP to text is also effortless.

With KP as an intermediate representation, a unified large-scale motion knowledge base is constructed.
Then, to fully exploit KP and the knowledge base, we build a motion understanding system with KP mediation. 
In detail, we learn a motion-KP joint latent space for multiple motion understanding applications, including motion interpolation, modification, and generation.
Moreover, leveraging the interpretability of KP, we propose a benchmark called Kinematic Prompts Generation (KPG), which generates motion from text prompts converted from KPs.
Thanks to the consistency and convenience of the KP-to-text conversion, KPG enables efficient white-box motion generation evaluation.
Given the bottle-necked evaluation of the current motion generation benchmarks~\cite{hml3d,kit} as shown in Fig.~\ref{fig:bench}, KPG could be a promising \textit{complementary} benchmark.
Results in Tab.~\ref{tab:phrase} show that KPG brings novel insights into the current text-to-motion progress.

Our contributions are:
(1) We propose KP as an intermediate representation to bridge the gap between motion and action semantics.
(2) We build a novel motion understanding system using KP and the aggregated large-scale knowledge base.
(3) We propose KPG as a benchmark for efficient white-box motion generation evaluation. 
Promising results are achieved on motion interpolation and generation tasks.
Moreover, extensive user studies are conducted, verifying the efficacy of our methods, and the consistency between KPG evaluation and human perception.
\section{Related Works}

\subsubsection{Motion Representation}
An intuitive motion representation is a sequence of static pose representations, like joint locations and limb rotations.
Efforts are paid to address the discontinuity of rotation for deep-learning methods~\cite{rot6d,bregier2021deep}.
Recent works on parametric body models~\cite{smpl,smplx} enable a more realistic body representation. 
Meanwhile, \cite{pons2014posebits} proposed Posebits, representing pose with boolean geometric part relationships.
\cite{posescript,delmas2023posefix} translates Posebits into text descriptions.
These abstract representations are flexible and insensitive to little perturbations, but their static nature ignores motion dynamics.
\cite{tang2022flag3d} acquire similar fine-grained descriptions from human annotation, while \cite{xiang2022language,SINC} adopted large-scale language models.
However, few recognize their potential in bridging the low-level motion and the high-level action semantics.
Phase functions~\cite{holden2020learned}, Labanotations~\cite{von1975laban}, and learned Motion Words~\cite{motionword} were also explored, though limited to specific actions like locomotion and dancing.

\subsubsection{Motion Generation} can be conditioned by its prefix/suffix~\cite{hernandez2019human,athanasiou2022teach,guo2023back}, action categories~\cite{actor,guo2020action2motion,ActFormer}, or audio~\cite{aist,li2021ai}.
Text-based motion generation has developed rapidly with the proposal of text-motion datasets~\cite{babel,hml3d}. 
Early works~\cite{temos,hml3d,Qian_2023_ICCV} used VAEs, while some~\cite{tevet2022motionclip,hong2022avatarclip,oohmg} extended the CLIP~\cite{clip} space to motion. 
Diffusion models~\cite{motiondiffuse,hmdm,Mofusion,Fg-T2M} were also adopted. 
MAA~\cite{maa} adopted a U-Net structure.
Some~\cite{ReMoDiffuse,Petrovich_2023_ICCV} explored retrieval-based methods.
\cite{Karunratanakul_2023_ICCV} aimed at controllable generation, while \cite{Yuan_2023_ICCV} introduced physical constraints.
However, most approaches still suffer from the gap between motion and action semantics.
Multiple methods~\cite{posegpt,guo2022tm2t,t2mgpt,mld,UDE,AttT2M,Kong_2023_ICCV} adopted (VQ-)VAE-compressed motion representation as mediation, while in the current data-limited situation, we identify that this single-modality compression might be sub-optimal.
Instead, KP could alleviate this by introducing explicit semantic-geometric correlation.

\section{Kinematic Phrase Base}
\subsection{Kinematic Phrases}
\label{sec:kinematic-phrase}
\begin{figure}[!t]
    \centering
    \includegraphics[width=.95\linewidth]{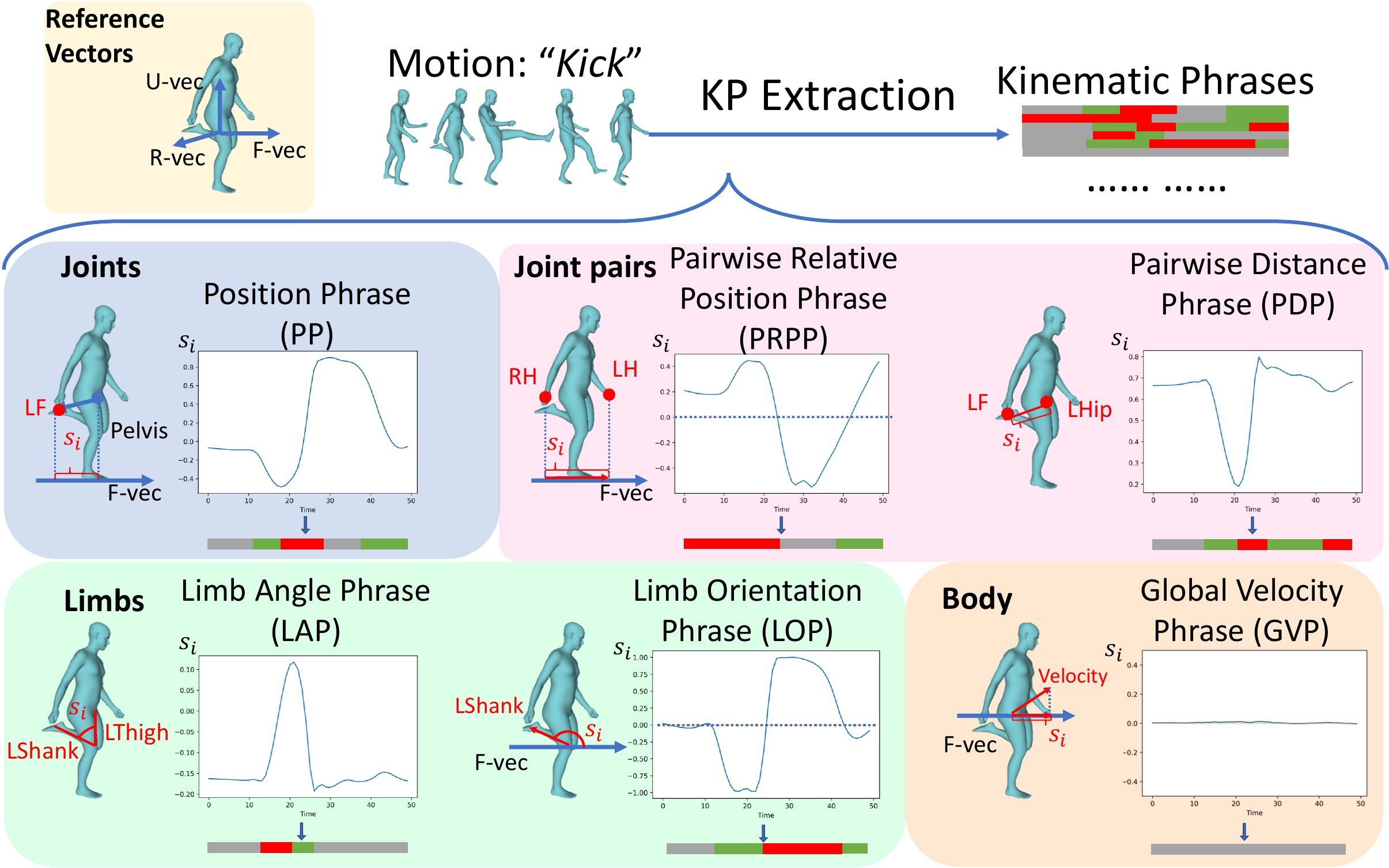}
    \caption{Six types of KP from four kinematic hierarchies are extracted from a motion sequence. A scalar indicator $s_i$ is calculated per Phrase \textit{per frame}. Its sign categorizes the corresponding Phrase.}
    \label{fig:2}
\end{figure}

Kinematic Phrases abstract motion into objective kinematic facts like \texttt{left-hand moves up} qualitatively.
We take inspiration from previous kinematic motion representations~\cite{von1975laban} and qualitative static pose representations~\cite{posescript,pons2014posebits,alphapose}, proposing six types of KP to comprehensively represent motion from different kinematic hierarchies:
For \textbf{joint movements}, there are 34 Position Phrases (PPs).
For \textbf{joint pair movements}, there are 242 Pairwise Relative Position Phrases (PRPPs) and 81 Pairwise Distance Phrases (PDPs).
For \textbf{limb movements}, there are 8 Limb Angle Phrases (LAPs) and 24 Limb Orientation Phrases (LOPs).
For \textbf{whole-body movements}, there are 3 Global Velocity Phrases (GVPs).
KP extraction is based on a skeleton sequence $X=\{x_i|x_i\in \mathcal{R}^{n_k\times 3}\}_{i=1}^t$, where $n_k$ is the number of joints ($n_k=17$ here), $x_i$ is the joint coordinates at $i$-th frame, and $t$ is the sequence length.
Note that $x_i^0$ indicates the pelvis/root joint.
For each Phrase, a scalar indicator sequence is calculated from the skeleton sequence.
Phrases are extracted as per-frame categorical representations w.r.t. indicator signs.
Unlike previous efforts~\cite{pons2014posebits,posescript}, we limit the KP criteria as the indicator signs to minimize human-defined biases (e.g., numerical criteria on the closeness of two joints) for objectivity and actuality.
Fig.~\ref{fig:2} illustrated the procedure.

{\bf Reference Vectors} are first constructed, indicating right, upward, and forward directions from a human \textit{cognitive view}.
We aim at the \textit{egocentric} reference frames that human tends to use when performing actions.
The negative direction of gravity is adopted as upward vector $r^u$, the vector from left hip to right hip is adopted as right vector $r^r$, and the forward vector is calculated as $r^f=r^u \times r^r$.
These vectors of each frame are denoted as $R^{\cdot}=\{r^{\cdot}_i\}_{i=1}^t$.

{\bf Position Phrase (PP)} focuses on the movement direction of joint $x^j$ w.r.t. reference vector $R^{\cdot}$.
The indicator for PP at $i$-th frame is calculated as
\begin{equation}
s_i^{(j,\cdot)}=\langle (x_i^j - x_i^0), r^{\cdot}_i\rangle - \langle (x_{i-1}^j - x_{i-1}^0), r^{\cdot}_{i-1}\rangle.
\end{equation}
The sign of $s_i^{(j,\cdot)}$ categorizes PP into moving \texttt{along/against} $R^{\cdot}$, or \texttt{relatively static} along $R^{\cdot}$ for indicators with small amplitudes.
12 joints (except the pelvis, eyes, and hips) and 3 reference vectors resulted in originally 36 PPs. Specifically, shoulders are excluded for the leftward vector, and hips are excluded for the upward vector, resulting in 36-2=34 PPs.

{\bf Pairwise Relative Position Phrase (PRPP)} describes the relative position of a pair of joints $(x^j, x^k)$  w.r.t. reference vector $R^{\cdot}$.
PRPP indicator at $i$-th frame is
\begin{equation}
s_i^{(j,k,\cdot)}=\langle (x_i^j - x_i^k), r^{\cdot}_i\rangle.
\end{equation}
For \texttt{(L-Hand, R-Hand)} and forward vector $R^f$, PRPP could be \texttt{L-Hand behind /in front of R-Hand} according to the sign of $s_i^{(j,k,\cdot)}$.
136 joint pairs and 3 reference vectors composed 408 Phrases originally. 24 joint pairs linked by limbs are filtered out. 30 Eye-related pairs are filtered out except (left eye, right eye) due to the others are covered by head-related pairs. 4 triplets that the relationship barely changes along the reference vector (e.g., right knee, left hip, and leftward vector) are filtered out. These result in 408-24*3-30*3-4=242 PRPPs.

{\bf Pairwise Distance Phrase (PDP)} describes how the L2 distance between a pair of joints $(x^j, x^k)$ changes. 
The indicator for PDP is calculated as
\begin{equation}
    s_i^{(j,k)}=\|x_i^j - x_i^k\|_2 - \|x_{i-1}^j - x_{i-1}^k\|_2.
\end{equation}
The sign of $s_i^{(j,k)}$ categorizes PDP into moving \texttt{closer/away}, or \texttt{relatively static}.
Among the original 105 joint pairs, 24 pairs that are in the skeleton topology are excluded, such as the hand and elbow, resulting in 81 PDPs.

{\bf Limb Angle Phrase (LAP)} targets at the change of bend angle between two connected limbs $(x^j,x^k)$ and $(x^j, x^l)$.
The LAP indicator is calculated as
\begin{equation}
    s_i^{(j,k,l)}=arccos(\langle x^k_i-x^j_i, x^l_i-x^j_i \rangle) - arccos(\langle x^k_{i-1}-x^j_{i-1}, x^l_{i-1}-x^j_{i-1} \rangle).
\end{equation}
8 LAPs describe the limb chain $(x^j,x^k)$-$(x^j, x^l)$ as \texttt{bending} or \texttt{unbending}.

{\bf Limb Orientation Phrase (LOP)} describes the orientation of the limb $(x^j,x^k)$ w.r.t. $R^{\cdot}$, with $x^k$ being the distal limb.
The LOP indicator is 
\begin{equation}
s_i^{(j,k,\cdot)}=\langle x^k_i-x^j_i, r^{\cdot}_i \rangle.
\end{equation}
The sign of $s_i^{(j,k,\cdot)}$ categorizes the LOP $(x^j,x^k)$ pointing \texttt{along/against $R^{\cdot}$}, or a placeholder category for those with little magnitude.
4 arms, 4 leg limbs, head, collarbones, hips, torsos, and upper body, paired with the 3 reference vectors, result in 45 LOPs in total. 21 pairs that barely change along the reference vector (e.g., left-right hip, leftward vector) are filtered out, resulting in 45-21=24 LOPs.

{\bf Global Velocity Phrase (GVP)} describes the direction of global velocity with respect to $R^{\cdot}$.
The indicator is calculated as 
\begin{equation}
    s_i^{\cdot}=\langle x^0_{i+1}-x^0_i, r^{\cdot}_i \rangle.
\end{equation}
The three categories are moving \texttt{along/against $R^{\cdot}$}, or \texttt{static along $R^{\cdot}$} according to the sign of $s_i^{\cdot}$.

These result in 392 Phrases in total, covering motion diversity and distribution from various levels.
While we clarify that these Phrases do not rule out the possibility of other possible useful potentials.

\subsection{Constructing Kinematic Phrase Base}
\label{sec:constructing-kinematic-phrase-base}
KP enables us to unify motion data with different formats to construct a large knowledge base containing motion, text, and KP thanks to its generality.
Motion sequences of different representations are collected, including 3D skeleton sequences and SMPL~\cite{smpl} /SMPL-X~\cite{smplx} parameter sequences.
The sequences are first re-sampled to 30Hz and rotated so that the negative direction of the z-axis is the gravity direction.
Then, the sequences are converted into 3D skeleton sequences for KP extraction as in Sec.~\ref{sec:kinematic-phrase}.
Text annotations attached to the sequences are directly saved.
For sequences with action category annotation, the category name is saved.
For those with neither text nor action category, the text information is set from its attached additional information, like objects for SAMP~\cite{samp}.
Finally, we collect 140K motion sequences from 11 datasets.
More details are included in the appendix.

\section{Applications of KP}

\subsection{Motion Understanding with KP}
An intuitive application of KP is to leverage it for better motion understanding. 
To achieve this, we first learn a motion-KP joint space, which exploits the clarity and interpretability of KP. 
Then, we adopt the learned latent space for multiple motion understanding tasks.
An overview is illustrated in Fig.~\ref{fig:3}.

\subsubsection{Preliminaries.}
Motion is represented as a pose sequence with $n$ frames as $M=\{m_i\}_{i=1}^n$.
In detail, SMPL~\cite{smpl} pose parameters are transformed to the 6D representation~\cite{rot6d}, then concatenated with the velocity of the root, resulting in a 147-dim representation per frame.
KP is represented by indicator signs as $C=\{c_i\}_{i=1}^n, c_i\in \{1,0,-1\}^{392}$.

\subsubsection{Joint Space Learning.}
The joint space is learned with Motion VAE $\{\mathcal{E}_m,\mathcal{D}_m\}$ and KP VAE $\{\mathcal{E}_p,\mathcal{D}_p\}$. 
All VAEs follow the transformer-based architecture from \cite{actor}.
Motion encoder $\mathcal{E}_m$ takes motion $M$ and two distribution tokens $m_{\mu},m_{\sigma}$ as input, and the outputs corresponding to the distribution tokens are taken as the $\mu_m$ and $\sigma_m$ of the Gaussian distribution.
Then, motion decoder $\mathcal{D}_m$ takes $z_m \sim \mathcal{G}(\mu_m, \sigma_m)$ as $K, V$, and a sinusoidal positional encoding of the expected duration as $Q$.
The output is fed into a linear layer to reconstruct motion $\hat{M}$.
KP VAE $\{\mathcal{E}_p,\mathcal{D}_p\}$ operates similarly, with the sign of $\mathcal{D}_p$ output as the predicted KP $\hat{C}$.
Notice that the $\mathcal{D}_m,\mathcal{D}_p$ could take arbitrary combinations of $z_m,z_p$ as input, outputting $\hat{M}_{\cdot},\hat{C}_{\cdot}$.
To learn the joint space, besides the conventional loss of reconstruction and KL divergence, we also align the two VAEs similarly to \cite{temos}.
In this way, we guide the learned motion latent space with the structure of KP.
Meanwhile, we randomly corrupt samples during training by setting a small portion of KP as 0.
The idea is that the overall representation should be robust with a small portion of KP unknown. 
Also, the missing Phrases should be recovered from existing Phrases.
This mines the correlation among Phrases and increases the robustness of the space.
More details are in the supplementary.

\begin{figure}[!t]
    \centering
    \includegraphics[width=.9\linewidth]{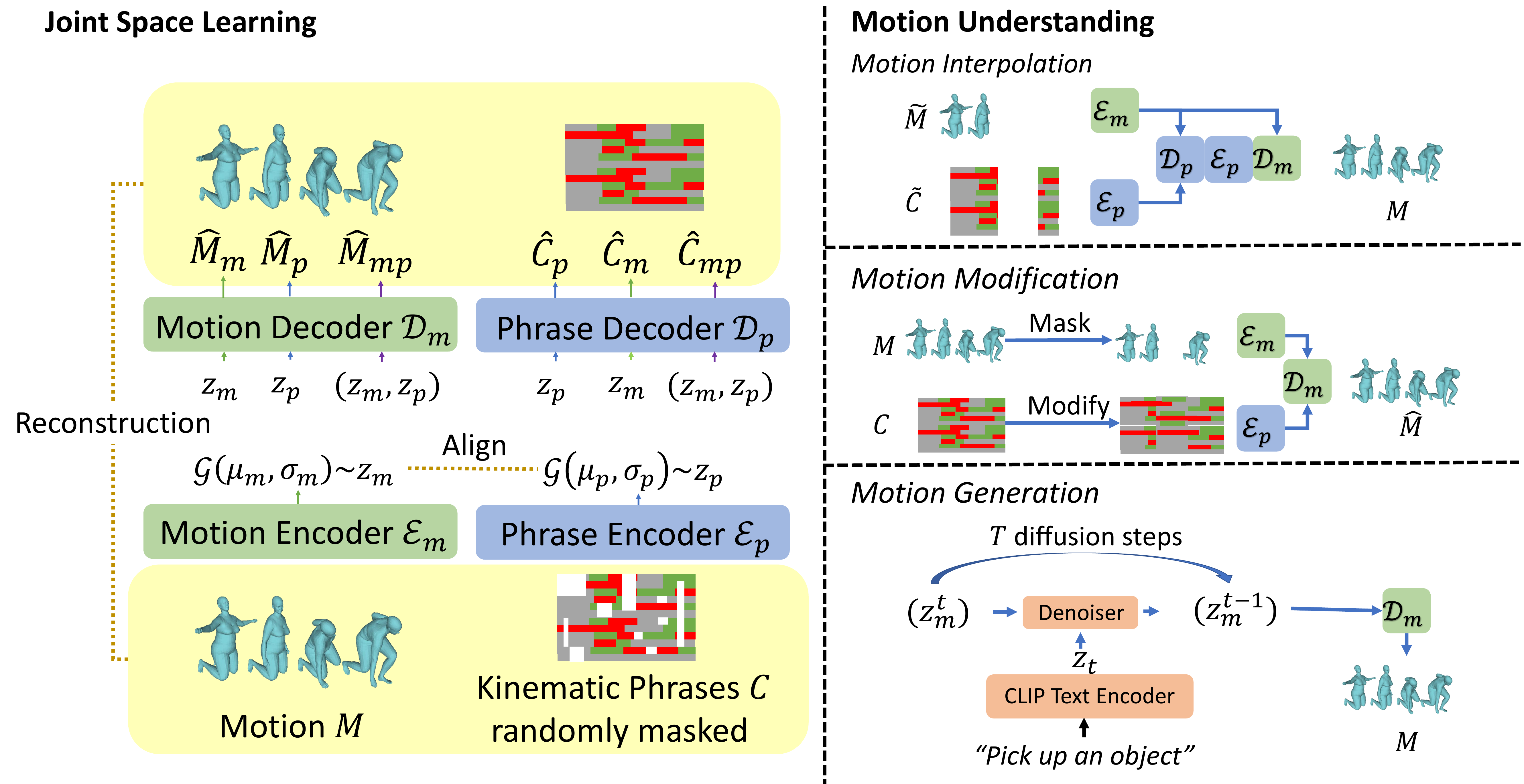}
    \caption{
    We train a motion-KP joint latent space to exploit the clarity and interpretability of KP.
    The space is then applied to multiple tasks, including motion interpolation, modification, and generation.
    }
    \label{fig:3}
\end{figure}

\subsubsection{Motion Understanding Tasks with Joint Space.}
\label{sec:bidirectional-mapping}
With the joint space, we can perform both low-level and high-level motion understanding tasks.
We introduce three applications to show the capability of KP, as shown in Fig.~\ref{fig:3}.
For \textbf{motion interpolation} given a corrupted motion sequence $\tilde{M}$, we extract its corresponding KP sequence $\tilde{C}$, then feed them to encoders $\mathcal{E}_m,\mathcal{E}_p$ and decoder $\mathcal{D}_p$, resulting in the estimated KP sequence $\hat{C}$.
$\hat{C}$ and $\tilde{M}$ are fed into $\mathcal{E}_m,\mathcal{E}_p$ and $\mathcal{D}_m$, resulting in interpolated $\hat{M}$.
For \textbf{motion modification} on motion $M$, modifications could be made on the extracted KP sequence $C$ resulting in $\tilde{C}$.
Motion frames to modify are then masked, getting $\tilde{M}$.
$\tilde{M},\tilde{C}$ are fed into $\mathcal{E}_m,\mathcal{E}_p$ and $\mathcal{D}_m$, getting the interpolated $\hat{M}$.
For \textbf{motion generation}, we simply replace the motion latent space in MLD~\cite{mld} with the latent space of our Motion VAE. 
Experiments show that KP manages to boost MLD with a considerable margin.
\subsection{Motion Benchmarking with Kinematic Prompt Generation}
\label{sec:kpg}
A more interesting and important application is benchmarking motion generation with the interpretability and objectivity of KP.

Before that, we first analyze current benchmarks. 
A crucial aspect of motion generation evaluation is motion-semantic consistency.
The gold standard is user study.
However, it is expensive and inefficient to scale.
Early metrics like MPJPE (Mean Per Joint Position Error) and MAE (Mean Angle Error) mechanically calculate the error between the generated and GT samples.
These metrics fail to reveal the real ability of generative models: 
What if the models memorize GT samples? Or what if the samples are diverse from GT but also true? 
FID (Frechet Inception Distance) is adopted to mitigate this issue. 
However, it provides a macro view of the quality of all generated samples without guarantees for individual samples. 
\cite{hml3d} proposed R-Precision, using a pre-trained text-motion matching model to examine whether the generated samples carry true semantics. 
They both rely on the latent space from a black-box pre-trained model, which is not credible.
Besides, models might learn short paths to over-fit the pre-trained model.
Meanwhile, since automatic mapping from motion to semantics is still an unsettled problem, adopting it to evaluate motion generation is not a decent choice. 
Moreover, most current motion generation evaluations are performed on datasets~\cite{hml3d,kit,uestc} with considerable complex everyday actions, further increasing the difficulty.
As shown in Fig.~\ref{fig:bench} and Tab.~\ref{tab:hml3d}, the current evaluation metrics are becoming increasingly indistinguishable.
Also, the higher-than-GT R-Precision could be insufficient as an indicator of semantic consistency.

\begin{figure}[!t]
    \centering
    \includegraphics[width=.9\linewidth]{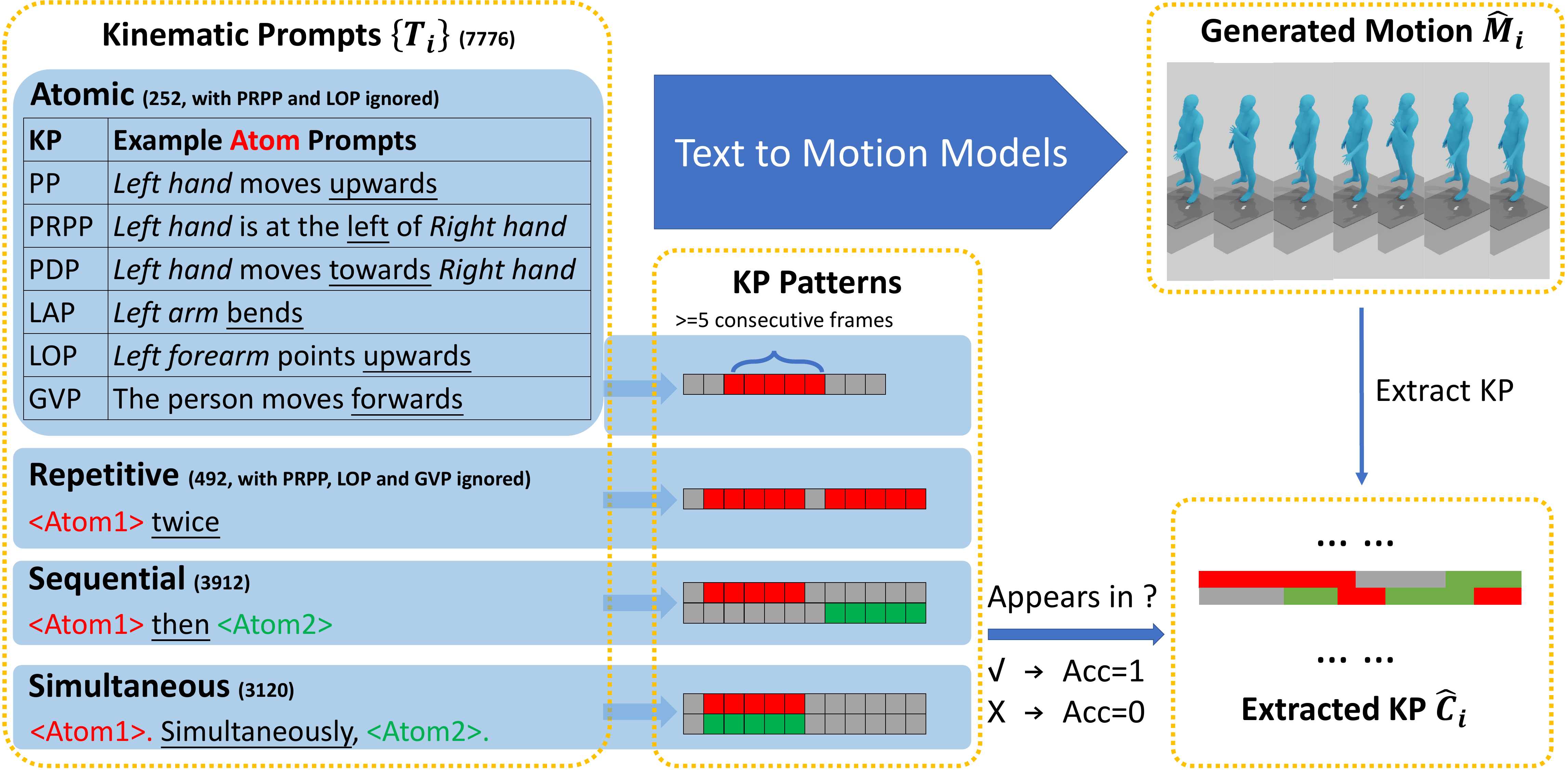}
    \caption{Kinematic Prompt Generation. 7,796 prompts are converted from KP with templates, corresponding to certain KP patterns. We calculate the generation accuracy by checking the appearance of the patterns in KP extracted from the generated motion.}
    \label{fig:kpg}
\end{figure}

To this end, we propose a novel benchmark: Kinematic Prompts Generation (KPG).
Instead of previous benchmarks focusing on everyday activities or sports, we step \textit{back} in the complexity of action semantics.
Based on KP, KPG focuses on evaluating \textit{whether the models could generate motion sequences consistent with specific kinematic facts given text prompts}.
An overview is given in Fig.~\ref{fig:kpg}.

In detail, we convert KP into 7,796 text prompts with templates as in Fig.~\ref{fig:kpg}.
The prompts could be categorized into 4 groups.
First, each \textbf{atomic prompt} involves only 1 KP, such as ``left hand moves upwards'', which evaluates the basic semantic consistency of text-to-motion models. Notice that PRPP and LOP are excluded since they represent static states, resulting in 252 prompts.
Second, each \textbf{repetitive prompt} involves a KP being repeated two or three times., like ``left arm bends twice'', evaluating the basic counting ability. PRPP, LOP, and GVP are excluded, bringing 492 prompts.
Third, each \textbf{sequential prompt} involves two independent KP to be sequentially executed, like ``left arm bends, then right arm bends''. 3,912 sequential prompts are randomly sampled.
Finally, each \textbf{simultaneous prompt} involves two independent KPs to be simultaneously executed, like ``left arm bends, and simultaneously, right arm bends.'' 3,120 prompts are randomly sampled.
The sequential and simultaneous prompts examine whether the model can understand the composition of simple actions.
Given prompt $T_i\in T$ from Phrase $c_i$ (or Phrases $(c_i^1,c_i^2)$), the model generates motion $\hat{M}_i$, along with extracted KP $\hat{C}_i$.
Then, whether $T_i$ is hit could be inferred via designed protocols by identifying certain KP patterns in $\hat{C}_i$.
In detail, the atomic prompt is identified as hit if $c_i$ appears for more than 5 consecutive frames in $\hat{C}_i$.
The repetitive prompt is identified as hit if and only if $c_i$ appears for at least 5 consecutive frames in $\hat{C}_i$ for corresponding times.
The sequential prompt is identified as hit if $c_i^2$ appears for at least 5 consecutive frames after (maybe not immediately) $c_i^1$ does the same.
The simultaneous prompt is identified as hit if $c_i^1$ and $c_i^2$ appear simultaneously for at least 5 consecutive frames.
The calculation involves no black-box model thanks to KP, presenting a fully white-box evaluation.
Also, with the effortless motion-to-KP conversion, the computation could be conducted automatically.
More details are in the appendix.

\section{Experiment}
\textbf{Implementation Details.}
HumanML3D~\cite{hml3d} test split is held out for evaluation, with the rest of KPB for training. 
The batch size is set as 288, and an AdamW optimizer with a learning rate of 1e-4 is adopted.
We randomly corrupt less than 20\% of the Phrases for a sample.
The Motion-KP joint space is trained for 6,000 epochs.
While the text-to-motion latent diffusion model is trained for 3,000 epochs, with the joint space frozen.
All experiments are conducted in 4 NVIDIA RTX 3090 GPUs.
More details are in the appendix.

\subsection{Motion Interpolation}
Following \cite{jiang2023motiongpt}, 50\% frames are randomly masked for interpolation.
In Tab.~\ref{tab:hml3d}, our method provides better FID.
With KPB, Diversity is increased.

\subsection{Motion Generation}
\label{sec:res-gen}

\subsubsection{Results on conventional text to motion} are shown in Tab.~\ref{tab:hml3d}.
We adopt the HumanML3D test set~\cite{hml3d} for conventional text-to-motion evaluation.
The evaluation model from \cite{hml3d} is adopted to calculate R-Precision, FID, Diversity, and Multimodality.
We run the evaluation 20 times and report the average value.
Our method is competitive without KPB.
However, there exist some unsatisfying observations.
First, KPB brings a counter-intuitive performance drop. 
Second, the metric values of most techniques are rather close (e.g., difference <0.03 for R-P@1), and no substantial advantage is perceived for one method over others.
Also, as discovered by MAA~\cite{maa}, despite its relatively poor R-Precision and FID, MAA manages to achieve better user reviews, indicating the potential loss of efficacy of these conventional metrics.
This is further revealed in Fig.~\ref{fig:rp_failure}, where a contradiction is observed between human judgment and R-precision.
Moreover, some methods show higher-than-GT R-P@1, which indicates that R-Precision might be insufficient as a semantic consistency indicator.
Given these, we further switch to KPG and pursue further insights.

\begin{table}[!t]
    \begin{minipage}{.6\linewidth}
    \centering
    \resizebox{\linewidth}{!}{\setlength{\tabcolsep}{0.8mm}{
    \begin{tabular}{lcccccc}
        \hline
                           & \multicolumn{2}{c}{Interpolation} & \multicolumn{4}{c}{Generation} \\
        Methods            & FID$\downarrow$ & Diversity$\to$ & R-P@1$\uparrow$ & FID$\downarrow$ & Diversity$\to$ & Multimodality \\
        \hline
        GT                 & 0.002 & 9.503 & 0.511 & 0.002 & 9.503 & - \\
        \hline
        TEMOS~\cite{temos}                  & - & -                   & 0.424 & 3.734 & 8.973 & 0.368 \\
        T2M~\cite{hml3d}                    & - & -                   & 0.455 & 1.067 & 9.188 & 2.090 \\
        MotionDiffuse~\cite{motiondiffuse}  & - & -                   & 0.491 & 0.630 & 9.410 & 0.730 \\
        MDM~\cite{hmdm}                     & 2.698 & 8.420           & 0.320 & 0.544 & 9.559 & 2.799 \\
        TM2T~\cite{guo2022tm2t}             & - & -                   & 0.424 & 1.501 & 8.589 & 2.424 \\
        MLD~\cite{mld}                      & - & -                   & 0.481 & 0.473 & 9.724 & 2.413 \\
        T2M-GPT~\cite{t2mgpt}               & - & -                   & 0.492 & 0.141 & 9.722 & 1.831 \\
        Fg-T2M~\cite{Fg-T2M}                & - & -                   & 0.492 & 0.243 & 9.278 & 1.614 \\
        M2DM~\cite{Kong_2023_ICCV}          & - & -                   & 0.497 & 0.352 & 9.926 & 3.587 \\
        AttT2M~\cite{AttT2M}                & - & -                   & 0.499 & 0.112 & 9.700 & 2.452 \\
        ReMoDiffuse~\cite{ReMoDiffuse}      & - & -                   & 0.510 & 0.103 & 9.018 & 1.795 \\
        MotionGPT~\cite{jiang2023motiongpt} & 0.214 & \textbf{9.560}  & 0.492 & 0.232 & \textbf{9.528} & 2.008 \\
        FineMoGen~\cite{zhang2024finemogen} & - & -                   & 0.504 & 0.151 & 9.263 & 2.696 \\
        GraphMotion~\cite{jin2024act}       & - & -                   & 0.504 & 0.116 & 9.692 & 2.766 \\
        AvatarGPT~\cite{zhou2023avatargpt}  & - & -                   & 0.510 & 0.168 & 9.624 & - \\
        GUESS~\cite{gao2024guess}           & - & -                   & 0.503 & 0.109 & 9.826 & 2.430 \\ \hline
        MDD~\cite{lou2023diversemotion}     & - & -                   & 0.515 & 0.072 & 9.683 & 1.869 \\
        MoMask~\cite{guo2023momask}         & - & -                   & \textbf{0.521} & \textbf{0.045} & - & 1.241 \\
        \hline
        Ours                                & \textbf{0.192} & 9.623  & 0.496 & 0.275 & 9.975  & 2.218 \\
        Ours w/ KPB                         & 0.214          & 10.142 & 0.471 & 0.641 & 10.262 & 2.891 \\
        \hline
    \end{tabular}}}
    \caption{Results on HumanML3D. R-P@1 is short for R-Precision@1. Most methods show comparable performance. Also, the higher-than-GT phenomenon indicates that R-Precision might be insufficient as a semantic consistency indicator.}
    \label{tab:hml3d}
    \end{minipage}
    \begin{minipage}{.4\linewidth}
        \centering
        \includegraphics[width=\linewidth]{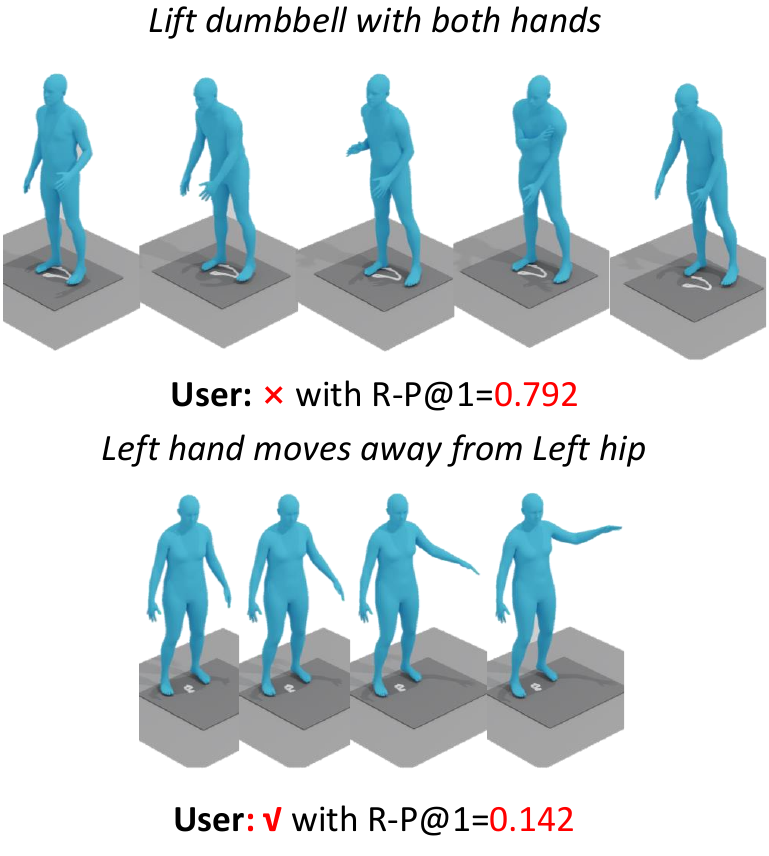}
        \captionof{figure}{R-Precision could contradict to user reviews, indicating that R-Precision could produce unfaithful evaluations of semantic consistency.}
        \label{fig:rp_failure}
    \end{minipage}
\end{table}

\subsubsection{Results on KPG} are shown in Tab.~\ref{tab:phrase}.
Besides accuracy, diversity and R-precision are reported.
We run the evaluation 20 times and report the average metric value.
The white-box designation of KPG evaluation makes it possible to evaluate the different aspects of the models.
Meanwhile, KPG is considered an easier task than conventional text-based motion generation since it is targeted at much less complex action semantics. 
Results show that previous methods are not performing as well as expected.
Also, R-Precision fails to model the performance on KPG with trivially low values.
Most methods perform well for atomic prompts, but when it comes to compositions, the accuracy drops.
Early efforts like HMDM~\cite{hmdm} and MotionDiffuse~\cite{motiondiffuse} are surprisingly competitive, even surpassing several more recent works.
T2M-GPT~\cite{t2mgpt} outperforms other methods by a considerable margin, but it fails for repetitive prompts.
An unexpected phenomenon is the failure of ReMoDiffuse~\cite{ReMoDiffuse}, which might be constrained by its retrieval-based methodology since most prompts in KPG are not available in the retrieval database.
Our method, which replaces the latent space of MLD~\cite{mld}, delivers substantial improvements, but the accuracy remains below 60\%, which is far from satisfying.
A considerable gap exists between existing efforts and ideal motion generation models.
More analyses are in the appendix.
\begin{table}[!t]
    \centering
    \resizebox{.8\linewidth}{!}{
    \begin{tabular}{lccccccc}
    \hline
                            & \multicolumn{5}{c}{Accuracy}                              & \multirow{2}{*}{Diversity} & \multirow{2}{*}{R-P@1}\\
    Methods                 & Atomic & Repetitive & Sequential & Simultaneous & Overall & & \\ \hline
    HMDM~\cite{hmdm}        & 93.25  & 23.98      & 56.11      & 30.90        & 45.16   & 5.923 & 0.034\\
    MotionDiffuse~\cite{motiondiffuse} & 93.21 & 22.15 & 51.54 & 27.76        & 41.49   & 5.257 & 0.051\\
    T2M-GPT~\cite{t2mgpt}   & \underline{97.22}  & 1.02       & \underline{70.24}      & \underline{42.47}        & \underline{55.59}   & 6.153 & 0.042\\
    MLD~\cite{mld}          & 92.06  & 22.76      & 48.26      & 28.01        & 39.94   & 5.526 & 0.061\\
    ReMoDiffuse~\cite{ReMoDiffuse} & 53.17  & 3.05       & 32.16      & 31.06        & 30.56   & 5.809 & 0.048 \\
    MotionGPT~\cite{jiang2023motiongpt}       & 92.06  & 16.87      & 55.98      & 31.76        & 44.96   & 5.776 & 0.039 \\
    MoMask~\cite{guo2023momask}          & 96.43  & \textbf{28.86} & 59.10      & 32.66        & 47.79   & 5.965 & 0.060 \\ \hline
    Ours                    & \textbf{98.80} & \underline{26.02} & \textbf{71.32} & \textbf{43.08}  & \textbf{57.86} & 6.048 & 0.053\\ \hline 
    \end{tabular}}
    \caption{Results on KPG demonstrate the various strengths of different methods.}
    \label{tab:phrase}
\end{table}

\subsubsection{User Study.} 
To reach a convincing performance comparison, we conduct user studies on both HumanML3D and KPG.
Our user study is different from previous efforts in two aspects.
First, instead of testing a small set of text prompts (less than 50 in previous works~\cite{hmdm,mld}), we randomly select 600 sentences from the HumanML3D test set.
By scaling up, the result is convincing in reflecting the ability to generate motion for diverse text inputs.
Second, without asking the volunteers to give a general rating for each sample or to choose between different samples, we ask them two questions: 1) Do the motion and the text match? and 2) Is the motion natural?
For Q1, three choices are given as ``No, Partially, Yes''.
For Q2, two choices are given as ``Yes, No''.
In this way, we explicitly evaluate two aspects of text-to-motion tasks: semantic consistency and naturalness, corresponding to R-Precision and FID.
For each prompt, we generate one sample considering the annotation cost.
We claim that the models should generate natural text-matching motion most of the time so that the one-sample setting would not hurt the fidelity of our user study.
36 volunteers are invited, each reviewing 200 sequences.
Thus each sequence receives 3 user reviews.
Also, we compute R-precision@1 of the generated sequences.

\begin{figure}[!t]
    \begin{minipage}{0.5\linewidth}
        \centering
        \includegraphics[width=.9\linewidth]{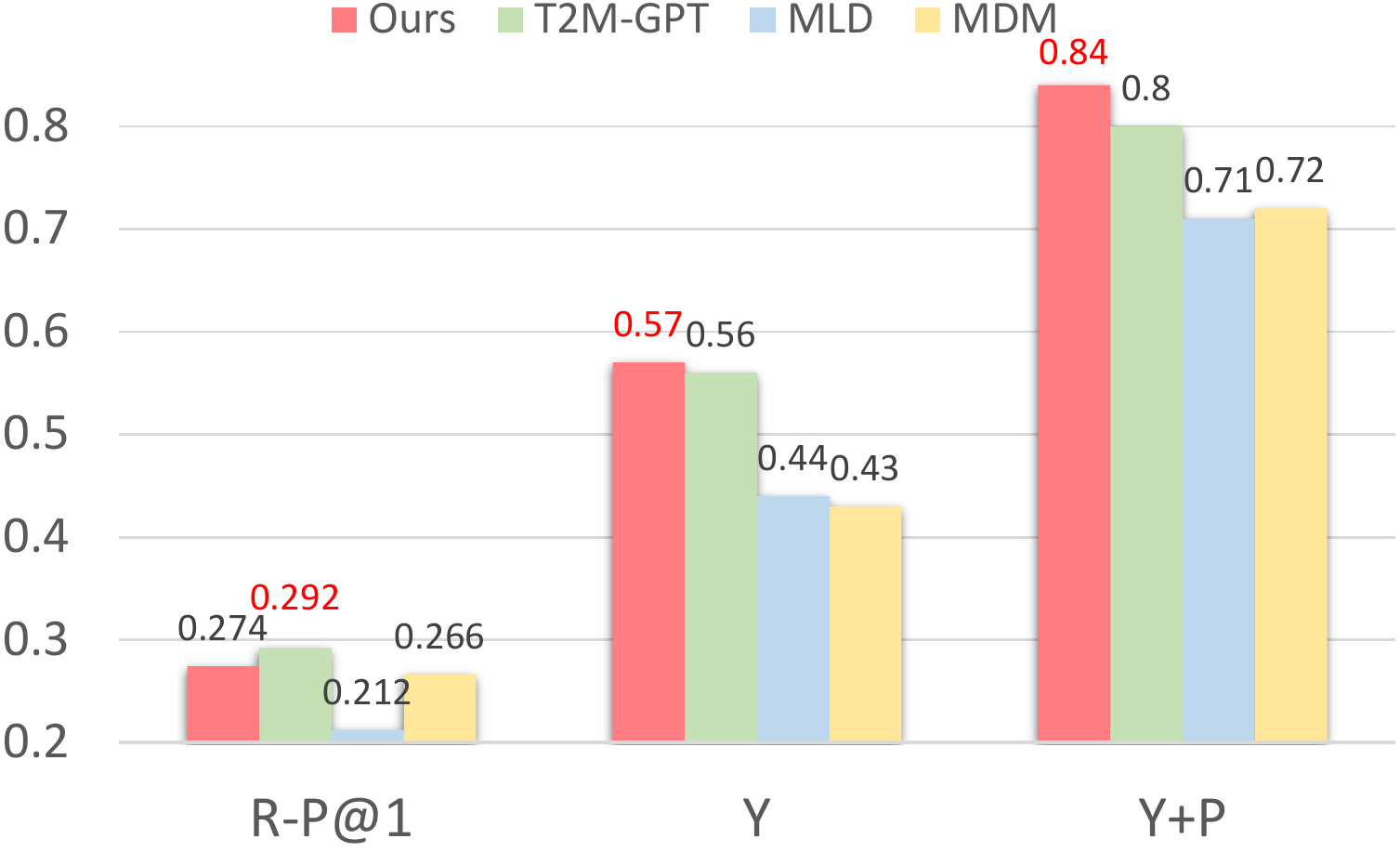}
        \captionof{figure}{
            User study on HumanML3D, with ``Y'' for Yes and ``P'' for partially.
        }
        \label{fig:hml3d_us}
    \end{minipage}
    \begin{minipage}{0.5\linewidth}
        \centering
        \includegraphics[width=.9\linewidth]{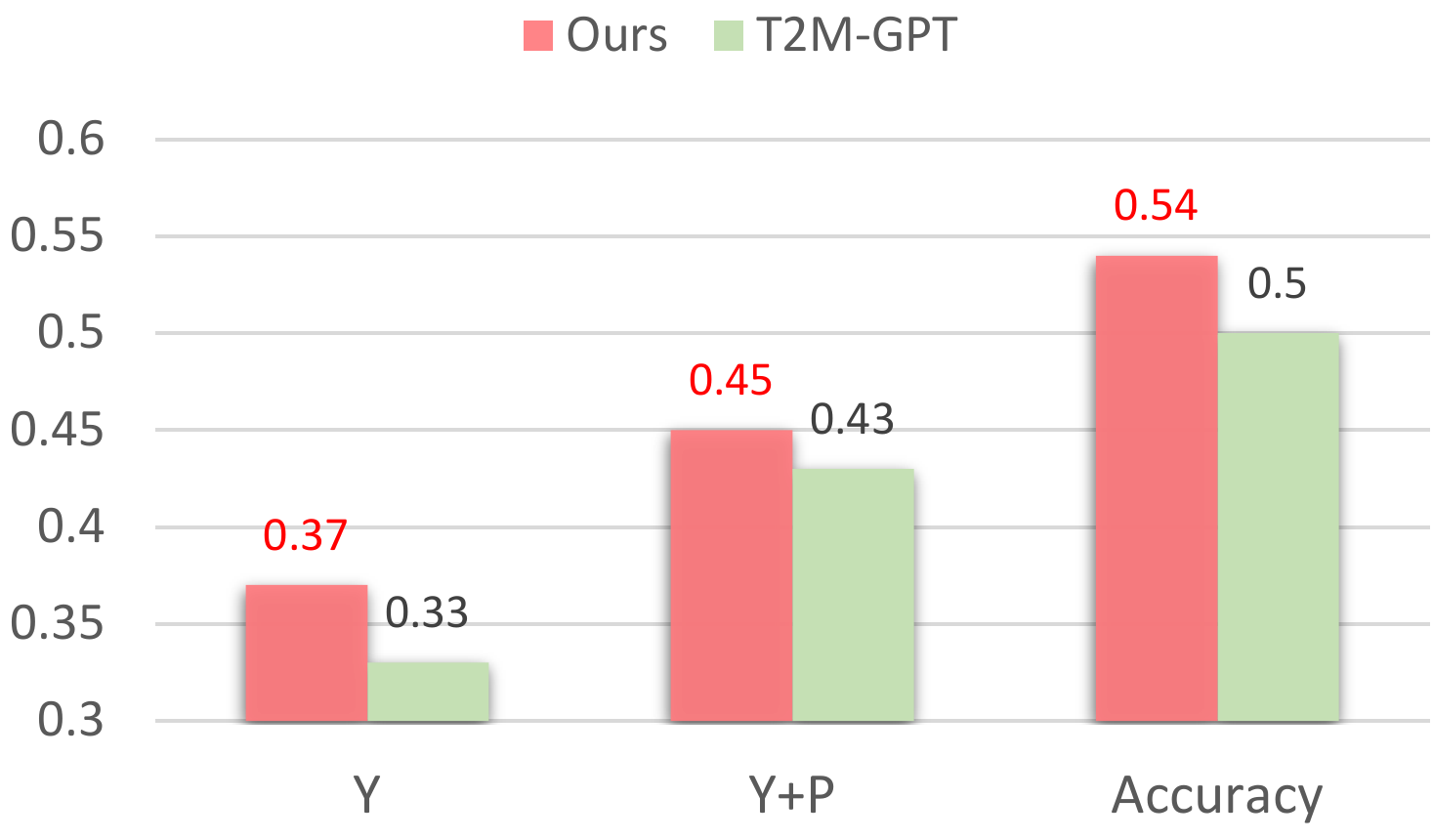}
        \captionof{figure}{User study on KPG, with ``Y'' for Yes and ``P'' for partially.}
        \label{fig:kpg_us}
    \end{minipage}
\end{figure}

Besides our model, we select MDM~\cite{hmdm} as a naive baseline, T2M-GPT~\cite{t2mgpt} as the SOTA model given the results in Tab.~\ref{tab:hml3d} and Tab.~\ref{tab:phrase}.
MLD~\cite{mld} is also selected as a direct baseline of our proposed method.
Results on HumanML3D~\cite{hml3d} are shown in Fig.~\ref{fig:hml3d_us}.
Though our method is not superior in R-Precision, we receive competitive user reviews, showcasing the efficacy of our motion-KP joint space. 
Also please notice that our model only contains \textbf{45.1M} parameters, while T2M-GPT contains \textbf{228M} parameters.
T2M-GPT and MLD present similar R-Precision, but only T2M-GPT manages to keep a good performance with user reviews.
Moreover, the discrepancy between R-Precision and user reviews is revealed in both absolute value and trends, also in Fig.~\ref{fig:rp_failure}.

A similar user study on KPG is conducted in Fig.~\ref{fig:hml3d_us} with 100 randomly selected prompts from KPG involving T2M-GPT and our model.
Fig.~\ref{fig:kpg_us} demonstrates that KP-inferred Accuracy and user reviews share similar trends.
Meanwhile, KP and user study give the same reviews for \textbf{84\%} of the samples.
We believe KPG could thus be a first step towards white-box automatic motion generation evaluation for semantic consistency.
More analyses are in the appendix.

\subsection{Visualization}
\label{sec:vis}
\begin{figure}[!t]
    \begin{minipage}{0.33\linewidth}
        \centering
        \includegraphics[width=.9\linewidth]{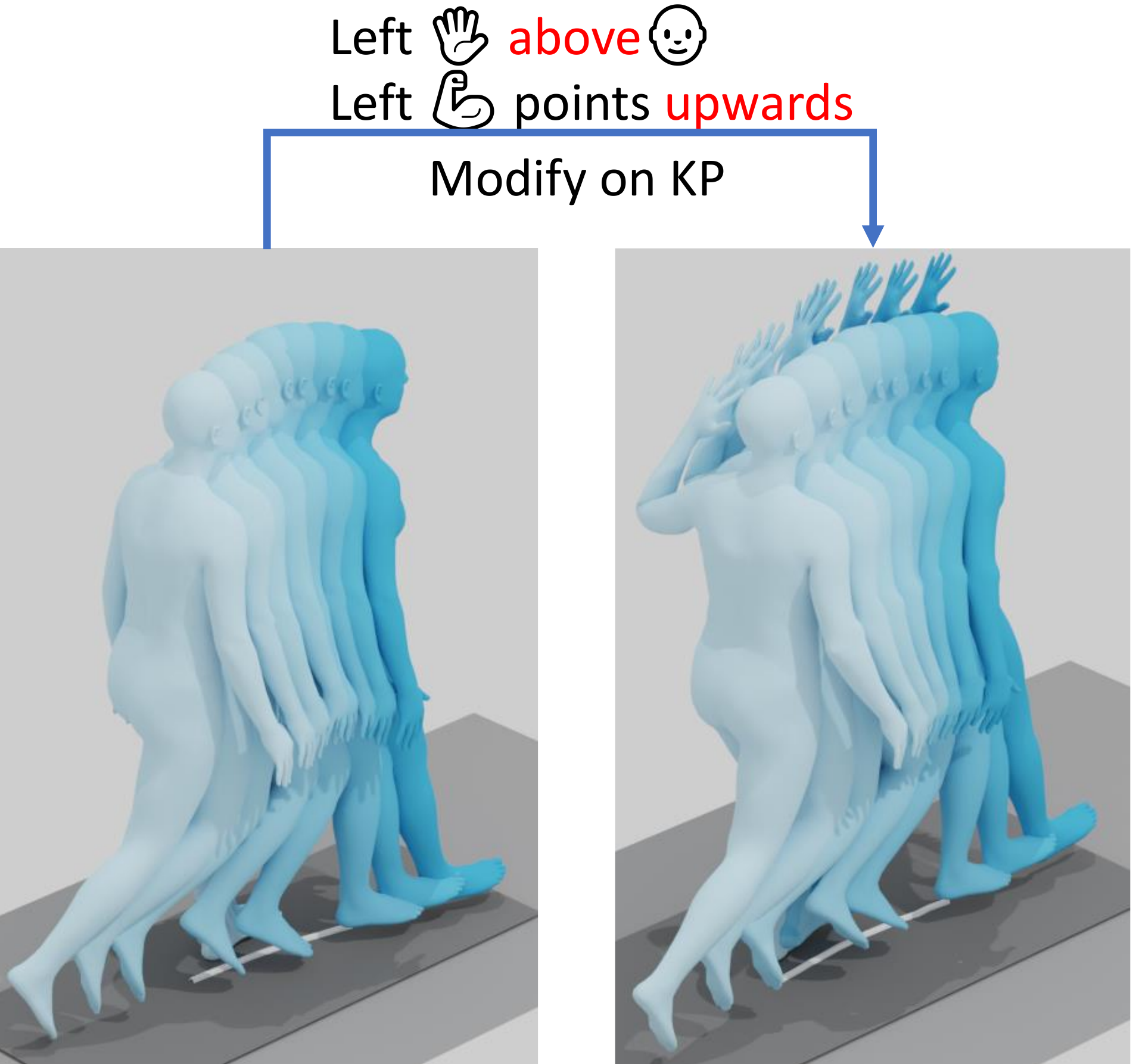}
        \captionof{figure}{Our model supports fine-grained modification on motion via KP.}
        \label{fig:modify}
    \end{minipage}
    \begin{minipage}{0.66\linewidth}
        \centering
        \includegraphics[width=.9\linewidth]{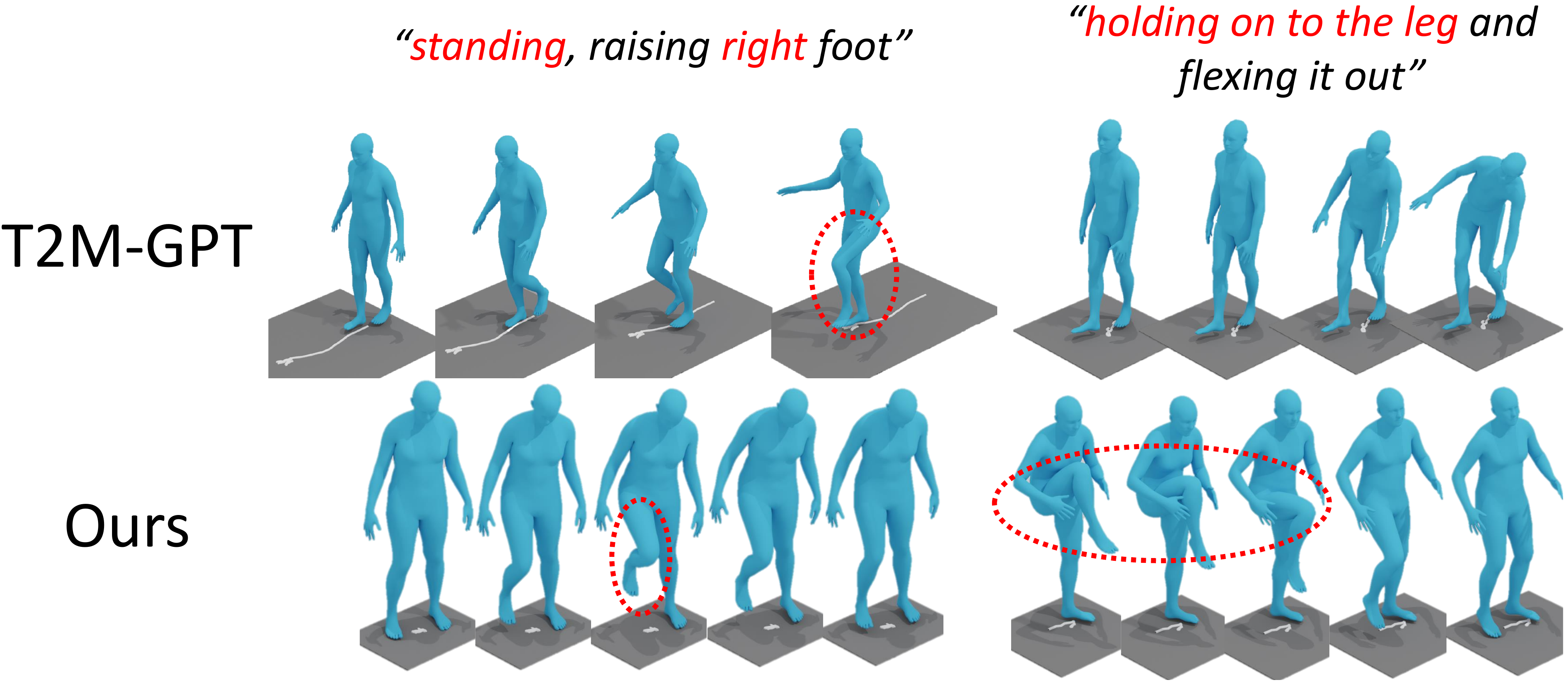}
        \captionof{figure}{Visualization of generated samples. Compared to T2M-GPT, our method provides a better response to prompts with explicit constraints on specific body parts.}
        \label{fig:vis}
    \end{minipage}
\end{figure}

A modification sample is shown in Fig.~\ref{fig:modify}.
By modifying KP, we could edit arbitrary motion at a fine-grained level.
Fig.~\ref{fig:vis} compares generated samples of T2M-GPT and our methods. 
Our method properly responds to text prompts with constraints on specific body parts thanks to KP, which explicitly models the action semantics with body part kinematics cues. 
Note that T2M-GPT generates redundant motion for simple prompts, while our method provides more concise and precise results.
More visualizations are in the appendix.

\subsection{Ablation Studies}
\label{sec:ablation}
KPG ablation study results are in Tab.~\ref{tab:ablation}.

\begin{wraptable}{R}{0.4\linewidth}
    \centering
    \resizebox{\linewidth}{!}{\setlength{\tabcolsep}{0.8mm}{
    \begin{tabular}{l c c}
        \hline
        Methods           & Acc.\%$\uparrow$ & Diversity  \\
        \hline
        Ours              & \textbf{57.86} & 6.048\\
        \hline
        w/o KP            & 39.94 & 5.526 \\
        \hline
        w/o Joint KP      & 50.03 & 5.685 \\
        w/o Joint Pair KP & 47.24 & 5.772 \\
        w/o Limb KP       & 55.92 & 5.934 \\
        w/o Body KP       & 56.84 & 5.871 \\
        \hline
    \end{tabular}}}
    \caption{KPG ablation results.}
    \label{tab:ablation}
\end{wraptable}

\textbf{KP latent.} By using a latent space without KP, a substantial performance drop is perceived, revealing the significance of KP in conveying action semantics.

\textbf{Different KP sets.} We examine the contribution of different KP sets: joint KP (PP), joint pair KP (PRPP, PDP), limb KP (LAP, LOP), and body KP (GVP).
A leave-one-out evaluation shows the elimination of joint KP and joint pair KP results in notable performance degradation, while the influence of the rest is relatively subtle.
\section{Discussion} 
\label{sec:discussion}
\subsubsection{Limitations.}
\textbf{First}, the current KP criteria guarantee objectivity but overlook kinematic information like amplitude and speed.
Also, due to the granularity of the adopted skeleton, fine-grained information on fingers is not well-preserved.
These would be promising for further exploration.
\textbf{Second}, KPB could be extended to datasets with other modalities, like 2D pose and egocentric action datasets.
Although these modalities provide incomplete 3D information, we could extract KP that is credibly accessible across modalities.
\textbf{Third}, with the convenient KP-to-text conversion, auxiliary motion descriptions could be automatically generated via KP. 
\textbf{Fourth}, KPG could be extended by paraphrasing existing prompts and combining different Phrases.
\textbf{Finally}, the proposed motion generation pipeline is devised for quick evaluation, while the motion-KP latent space could be incorporated with more advanced algorithms for better exploitation.

\subsubsection{Broader Impacts.} The proposed KP and KPG provide a white-box instrument to diagnose text-to-motion algorithms, which could help develop reliable motion tools. Better motion synthesis could advance fields including animation production, sports training, human-robot interaction, and rehabilitation. However, it could also be used for illegal media content or false information.

\section{Conclusion}
In this paper, we proposed an intermediate representation to bridge human motion and action semantics as the Kinematic Phrase.
KP achieved proper abstraction, interpretability, and generality by focusing on objective kinematic facts of human motion.
A motion understanding system based on KP was proposed and proven effective in motion interpolation, modification, and generation.
Moreover, Kinematic Prompt Generation is proposed as a white-box text-to-motion benchmark, bringing new insights to the literature.
We believe that KP has great potential to advance motion understanding.

\section*{Acknowledgments}
This work is supported in part by the
National Natural Science Foundation of China under Grants No.62306175, No.62302296, 
National Key Research and Development Project of China (No.2022ZD0160102, No.2021ZD0110704), 
Shanghai Artificial Intelligence Laboratory, XPLORER PRIZE grants.

%
%
\bibliographystyle{splncs04}
\bibliography{egbib}

\appendix
\section{Kinematic Phrase Details}
\label{sec:supp_kpd}
This section lists the details of the six defined types of KP.
During extraction, the indicator is set as zero if it is smaller than 1e-4.

\subsection{Position Phrase} 
There are 34 phrases, corresponding to 34 interested $\langle joint,\ reference\ vector \rangle$ pairs like $\langle left\ hand,\ forward\ vector \rangle$.
The pairs are listed in the file \texttt{KP/pp.txt}.

\subsection{Pairwise Relative Position Phrase} 
There are 242 phrases corresponding to 242 interested $\langle joint,\ joint, reference\ vector \rangle$ triplets like $\langle left\ hand,\ right\ hand, forward\ vector \rangle$, listed in the file \texttt{KP/prpp.txt}.

\subsection{Pairwise Distance Phrase} 
Joint pairs that are connected by human body topology are filtered out, like hand-elbow and shoulder-hip. 
There are 81 phrases corresponding to 81 interested $\langle joint,\ joint \rangle$ pairs like $\langle left\ hand,\ right\ hand \rangle$, listed in the file \texttt{KP/pdp.txt}.

\subsection{Limb Angle Phrase} 
There are 8 phrases corresponding to 8 interested limbs, listed in the file \texttt{KP/lap.txt}.

\subsection{Limb Orientation Phrase}
There are 24 phrases corresponding to 24 interested $\langle limb,\ reference\ vector \rangle$ pairs like $\langle left\ shank,\ right\ vector \rangle$, listed in the file \texttt{KP/lop.txt}. 

\subsection{Global Velocity Phrase}
There are 3 phrases corresponding to the velocity direction concerning the three reference vectors.

\section{Kinematic Phrase Base Details}
\label{sec:supp_kpbd}
\begin{table}[!t]
    \centering
    \begin{tabular}{l c c c c}
        \hline
        Dataset                 & Mot. Rep. & \#Seqs & \#Actions & Text \\
        \hline
        AMASS~\cite{amass}      & SMPL-X   & 26k  & 260 & \checkmark
        \\
        GRAB~\cite{grab}        & SMPL-X   & 1k   &  4  & \checkmark \\
        SAMP~\cite{samp}        & SMPL-X   & 0.2k & N/A & \checkmark*\\
        Fit3D~\cite{fit3d}      & SMPL-X   & 0.4k & 29  & \checkmark \\
        CHI3D~\cite{chi3d}      & SMPL-X   & 0.4k & 8   & \checkmark \\
        UESTC~\cite{uestc}      & SMPL     & 26k  & 40  & \checkmark \\
        AIST++~\cite{aist}      & SMPL     & 1k   & N/A & \checkmark*\\
        BEHAVE~\cite{behave}    & SMPL     & 0.3k & N/A & \checkmark*\\
        HuMMan~\cite{humman}    & SMPL     & 0.3k & 339 & \checkmark \\
        GTAHuman~\cite{gtahu}   & SMPL     & 20k  & N/A &  \text{\sffamily x}   \\
        Motion-X\cite{lin2023motion}& SMPL-X & 65k & N/A & \checkmark \\
        \hline
        \textbf{Sum}            & -        & \textbf{140k} & \textbf{680+} & - \\
        \hline
    \end{tabular}
    \caption{Statistics of Kinematic Phrase Base. \textit{Mot. Rep.} indicates motion representation. ``\checkmark*'' means texts are generated from the attached additional information instead of human annotation.}
    \label{tab:stat}
\end{table}

\begin{figure}[!t]
    \centering
    \includegraphics[width=\linewidth]{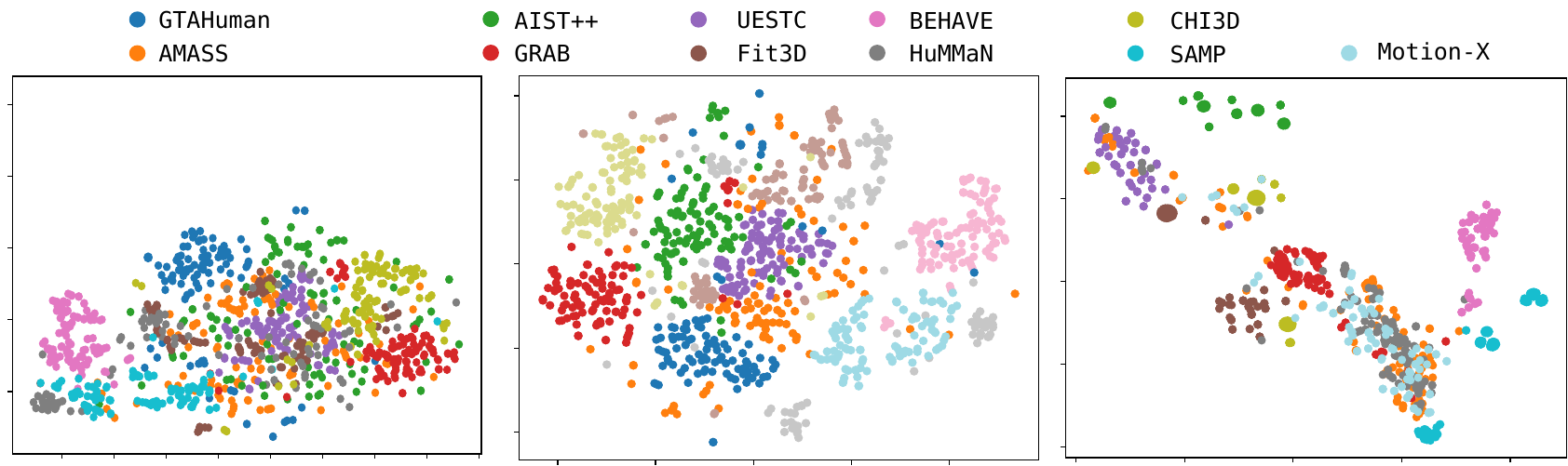}
    \caption{Motion, KP, and text distribution of Kinematic Phrase Base.}
    \label{fig:kpb-vis}
\end{figure}

\begin{figure}[!t]
    \centering
    \includegraphics[width=.7\linewidth]{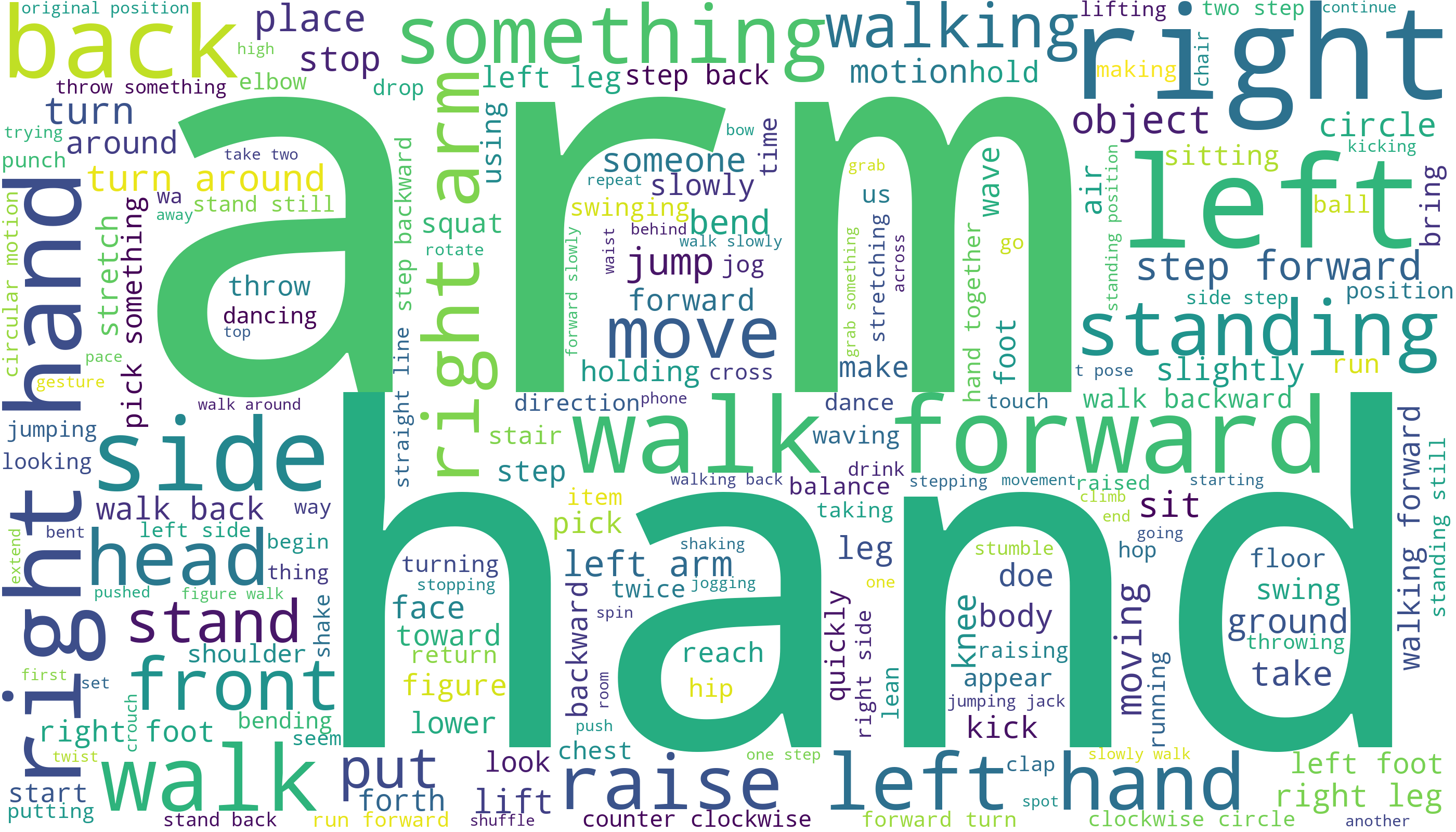}
    \caption{Word cloud visualization of the texts in Kinematic Phrase Base.}
    \label{fig:word-cloud}
\end{figure}

As shown in Tab.~\ref{tab:stat}, over \textbf{140 K} motion sequences are collected to construct the Kinematic Phrase Base, including \textbf{9 M} frames (in 30 FPS) with \textbf{48 K} different sentences, covering a vocabulary size of \textbf{7,418}.
Here, we illustrate the distribution of the collected database represented in motion, KP, and text in Fig.~\ref{fig:kpb-vis}.
Besides, a word cloud visualization of the texts in the database is illustrated in Fig.~\ref{fig:word-cloud}.

\section{Method Details}
\label{sec:supp_method}
\subsection{Losses for Joint Space Learning}
Reconstruction loss $\mathcal{L}_{rec}$ compares the GT with the outputs of the VAEs.
L1 losses are calculated for the motion representation $M, \hat{M}$, KP $C,\hat{C}$, the skeleton joints $J, \hat{J}$, the down-sampled mesh vertices $V,\hat{V}$, and the joint accelerations $A,\hat{A}$:
\begin{equation}
    \mathcal{L}_{rec}=\sum_{\cdot \in \{m,p,mp\}}||M_{\cdot}-\hat{M}_{\cdot}||_1  + ||C_{\cdot}-\hat{C}_{\cdot}||_1 + ||J_{\cdot}-\hat{J}_{\cdot}||_1 + ||V_{\cdot}-\hat{V}_{\cdot}||_1 + ||A_{\cdot}-\hat{A}_{\cdot}||_1. 
\end{equation}
KL divergence loss $\mathcal{L}_{KL}$ encourages each distribution to be similar to a normal distribution $\pi=\mathcal{G}(0, I)$ by minimizing the Kullback-Leibler(KL) divergence between the normal distribution and the learned motion and KP distributions.
The loss is calculated as 
\begin{equation}
    \mathcal{L}_{KL}=KL(\phi_m, \pi) + KL(\phi_p,\pi).
\end{equation}
Distribution alignment loss $\mathcal{L}_{da}$ encourages the distributions of motion and KP to resemble each other by minimizing the KL divergence between them.
The loss is calculated as 
\begin{equation}
    \mathcal{L}_{da}=KL(\phi_m, \phi_p) + KL(\phi_p,\phi_m).
\end{equation}
Embedding alignment loss $\mathcal{L}_{emb}$ encourages the sampled latent vectors to be aligned by minimizing their L1 distance.
The loss is calculated as 
\begin{equation}
    \mathcal{L}_{emb}=||z_m-z_p||_1.
\end{equation}
The overall loss is calculated as 
\begin{equation}
    \mathcal{L}=\lambda_1\mathcal{L}_{rec}+\lambda_2\mathcal{L}_{KL}+\lambda_3\mathcal{L}_{da}+\lambda_4\mathcal{L}_{emb}, 
\end{equation}
where $\{\lambda_i\}_{i=1}^4$ are weighting coefficients.

\section{Kinematic Prompt Generation Details}
\label{sec:supp_kpgd}
\subsection{Prompts}
We provide the 7,776 text prompts converted from KP in the file \texttt{kpg.txt}.

\section{Experiment Details}
\label{sec:supp_ed}
\subsection{Implementation Details}
Sequences are sampled to 15 FPS and randomly clipped into short clips with lengths between 30 frames and 150 frames.
The Motion VAE and KP VAE share the same structure: a 4-layer transformer encoder, a 4-layer transformer decoder, and a fully connected layer for final outputs.
The denoiser adopted for text-to-motion is designed as a 4-layer transformer decoder.
The latent size is set to 256.
$\{\lambda_i\}_{i=1}^4$ are set as 1.
The learning rate is decayed at 4,000 epochs for joint space training and at 2,000 epochs for text-to-motion latent diffusion model training.

\subsection{Motion Generation Settings}
For HumanML3D~\cite{hml3d}, motion sequences are generated for 10 seconds given a text prompt.
For KPG, the models are required to generate 120 frames given a text prompt.

R-Precision is calculated similarly to \cite{hml3d}.
For each generated motion, its text description is mixed with 31 randomly selected mismatched descriptions from the test set.
The cosine distances between the motion feature and text features are computed.
The average accuracy at the top-1 place is reported.

FID is adopted to measure the divergence between the GT motion distribution and the generated motion distribution in the latent space.

Diversity measures the variance of the generated motion sequences. It is calculated as the average latent distance between two randomly sampled generated motion sets. The set size is set as 300 in this paper.

Multimodality measures the variance of the generated motion sequences within each text prompt.
For each description, two subsets of motion sequences with the same size are generated, and then the Multimodality is calculated as the average distance between the two sets of motions in the latent space.
The size of each subset is set as 10 in this paper.

\subsection{Model Size comparison.}
We compare the number of parameters in our model and previous SOTAs in Tab.~\ref{tab:size_comp}.
As shown, with a model size comparable to MLD~\cite{mld} and significantly lower than T2M-GPT~\cite{t2mgpt}, we achieve competitive performance on conventional benchmarks and even better performance with the newly proposed KPG.
\begin{table}[!t]
    \centering
    \resizebox{.6\linewidth}{!}{
    \begin{tabular}{c|c|c|c|c}
        \hline
        Method   & MDM~\cite{hmdm} & MLD~\cite{mld} & T2M-GPT~\cite{t2mgpt} & Ours  \\ \hline
        \#params & 23M & 42.7M & 228M & 45.1M \\
        \hline
    \end{tabular}}
    \caption{Model Size Comparison.}
    \label{tab:size_comp}
\end{table}

\subsection{User Study Details}

\subsubsection{User Study Design}
As stated in the main text, we adopt a direct Q\&A-style user study instead of a popular preference test or ratings.
Here we clarify the reason for this design choice.
First, this design is more suitable in evaluating \textbf{semantic consistency}, which we identify as categorical instead of continuous at the sample level.
That is, it is hard to tell whether a motion is more \texttt{raising left-hand up} than another.
Instead, there is only whether a motion is \texttt{raising left-hand up} or not.
Therefore, we chose to present a direct question on semantic consistency.
Second, this design explicitly decouples the evaluation of text-to-motion into semantic consistency and naturalness, corresponding to R-Precision and FID.
When rating motions or choosing between two motions, it is hard to guarantee the users make choices according to the expected standard.
Therefore, we explicitly ask decoupled binary questions for decomposition.
Third, it helps reduce annotation costs.
For preference testing, the complexity is $O(N^2)$, while with our user-study protocol, the complexity is only $O(N)$.
In consideration of our primary focus on semantic consistency, we adopt this protocol.
We also admit this protocol is sub-optimal in naturalness evaluation, which is a continuous factor. 
We present the results on naturalness as a reference in the following sections.

\subsubsection{User Study on Conventional Text-to-Motion}
\begin{table}[t]
    \begin{subtable}[t]{0.49\linewidth}
        \resizebox{\linewidth}{!}{\begin{tabular}{cccccc}
        \hline
        \multicolumn{2}{c}{\multirow{2}{*}{\begin{tabular}[c]{@{}c@{}}FID = 0.544\\ R-P@1 = 0.266\end{tabular}}} &
          \multicolumn{3}{c}{Semantic consistency} &
          \multirow{2}{*}{Sum} \\
        \multicolumn{2}{c}{}               & Yes  & Partially & No   &      \\ \hline
        \multirow{2}{*}{Naturalness} & Yes & 0.40 & 0.18      & 0.10 & 0.68 \\
                                     & No  & 0.03 & 0.11      & 0.18 & 0.32 \\ \hline
        \multicolumn{2}{c}{Sum}            & 0.43 & 0.29      & 0.28 & 1    \\ \hline
        \end{tabular}}
        \caption{MDM~\cite{hmdm}.}
    \end{subtable}
    \begin{subtable}[t]{0.49\linewidth}
        \resizebox{\linewidth}{!}{\begin{tabular}{cccccc}
        \hline
        \multicolumn{2}{c}{\multirow{2}{*}{\begin{tabular}[c]{@{}c@{}}FID = 0.212\\ R-P@1 = 0.473\end{tabular}}} &
          \multicolumn{3}{c}{Semantic consistency} &
          \multirow{2}{*}{Sum} \\
        \multicolumn{2}{c}{}               & Yes  & Partially & No   &      \\ \hline
        \multirow{2}{*}{Naturalness} & Yes & 0.34 & 0.13      & 0.04 & 0.51 \\
                                     & No  & 0.10 & 0.14      & 0.25 & 0.49 \\ \hline
        \multicolumn{2}{c}{Sum}            & 0.44 & 0.27      & 0.29 & 1    \\ \hline
        \end{tabular}}
        \caption{MLD~\cite{mld}.}
    \end{subtable}

    \begin{subtable}[t]{0.49\linewidth}
        \resizebox{\linewidth}{!}{\begin{tabular}{cccccc}
        \hline
        \multicolumn{2}{c}{\multirow{2}{*}{\begin{tabular}[c]{@{}c@{}}FID = 0.141\\ R-P@1 = 0.292\end{tabular}}} &
          \multicolumn{3}{c}{Semantic consistency} &
          \multirow{2}{*}{Sum} \\
        \multicolumn{2}{c}{}               & Yes  & Partially & No   &      \\ \hline
        \multirow{2}{*}{Naturalness} & Yes & 0.50 & 0.16      & 0.05 & 0.71 \\
                                     & No  & 0.06 & 0.08      & 0.15 & 0.29 \\ \hline
        \multicolumn{2}{c}{Sum}            & 0.56 & 0.24      & 0.20 & 1    \\ \hline
        \end{tabular}}
        \caption{T2M-GPT~\cite{t2mgpt}.}
    \end{subtable}
    \begin{subtable}[t]{0.49\linewidth}
        \resizebox{\linewidth}{!}{\begin{tabular}{cccccc}
        \hline
        \multicolumn{2}{c}{\multirow{2}{*}{\begin{tabular}[c]{@{}c@{}}FID = 0.631\\ R-P@1 = 0.274\end{tabular}}} &
          \multicolumn{3}{c}{Semantic consistency} &
          \multirow{2}{*}{Sum} \\
        \multicolumn{2}{c}{}               & Yes  & Partially & No   &      \\ \hline
        \multirow{2}{*}{Naturalness} & Yes & 0.52 & 0.21      & 0.02 & \textbf{0.75} \\
                                     & No  & 0.05 & 0.06      & 0.14 & 0.25 \\ \hline
        \multicolumn{2}{c}{Sum}            & \textbf{0.57} & \textbf{0.27}      & 0.16 & 1    \\ \hline
        \end{tabular}}
        \caption{Ours.}
    \end{subtable}
    \caption{Detailed user study results on HumanML3D.}
    \label{tab:detail_us_hml3d}
\end{table}

Detailed results of the HumanML3D user study are demonstrated in Tab.~\ref{tab:detail_us_hml3d}.
As shown, both FID and R-P@1 are not consistent with the user reviews, indicating these black-box-based metrics might be sub-optimal for motion generation evaluation.
Meanwhile, the four evaluated methods present a similar positive correlation between semantic consistency and naturalness.
Moreover, it shows that generating natural motions is a little harder than generating partially semantic-consistent motions, which might be a potential direction to advance motion generation.

\subsubsection{User Study on KPG}

\begin{table}[t]
    \begin{subtable}[t]{0.49\linewidth}
        \resizebox{\linewidth}{!}{\begin{tabular}{cccccc}
        \hline
        \multicolumn{2}{c}{\multirow{2}{*}{Accuracy = 50\%}} & \multicolumn{3}{c}{Semantic consistency} & \multirow{2}{*}{Sum} \\
        \multicolumn{2}{c}{}               & Yes  & Partially & No   &      \\ \hline
        \multirow{2}{*}{Naturalness} & Yes & 0.29 & 0.09      & 0.53 & 0.91 \\
                                     & No  & 0.04 & 0.01      & 0.04 & 0.09 \\ \hline
        \multicolumn{2}{c}{Sum}            & 0.33 & \textbf{0.10}      & 0.57 & 1    \\ \hline
        \end{tabular}}
        \caption{T2M-GPT~\cite{t2mgpt}.}
    \end{subtable}
    \begin{subtable}[t]{0.49\linewidth}
        \resizebox{\linewidth}{!}{\begin{tabular}{cccccc}
        \hline
        \multicolumn{2}{c}{\multirow{2}{*}{Accuracy = \textbf{54\%}}} &
          \multicolumn{3}{c}{Semantic consistency} &
          \multirow{2}{*}{Sum} \\
        \multicolumn{2}{c}{}               & Yes  & Partially & No   &      \\ \hline
        \multirow{2}{*}{Naturalness} & Yes & 0.33 & 0.07      & 0.51 & \textbf{0.92} \\
                                     & No  & 0.04 & 0.01      & 0.04 & 0.08 \\ \hline
        \multicolumn{2}{c}{Sum}            & \textbf{0.37} & 0.08      & 0.55 & 1    \\ \hline
        \end{tabular}}
        \caption{Ours.}
    \end{subtable}
    \caption{Detailed user study results on KPG.}
    \label{tab:detail_us_kpg}
\end{table}

Detailed user study results on KPG are demonstrated in Tab.~\ref{tab:detail_us_kpg}. 
Our proposed Accuracy shares a similar trend with user-reviewed semantic consistency between the two methods. 
Both methods receive good naturalness reviews, which could result from the simple prompt structure of KPG.

\begin{table}[t]
    \centering
    \begin{tabular}{cccccc}
    \hline
    \multicolumn{2}{c}{\multirow{2}{*}{}} & \multicolumn{3}{c}{User Reviewed} & \multirow{2}{*}{Sum} \\
    \multicolumn{2}{c}{}               & Yes  & Partially & No   &      \\ \hline
    \multirow{2}{*}{KP-Inferred} & Yes & 0.32 & 0.08      & 0.12 & 0.52 \\
                                 & No  & 0.03 & 0.01      & 0.44 & 0.48 \\ \hline
    \multicolumn{2}{c}{Sum}            & 0.35 & 0.09      & 0.56 & 1    \\ \hline
    \end{tabular}
    \caption{Detailed consistency statistics between KP-inferred Accuracy and user-reviewed semantic consistency.}
    \label{tab:kpg-consistency}
\end{table}

\begin{figure}
    \centering
    \includegraphics[width=\linewidth]{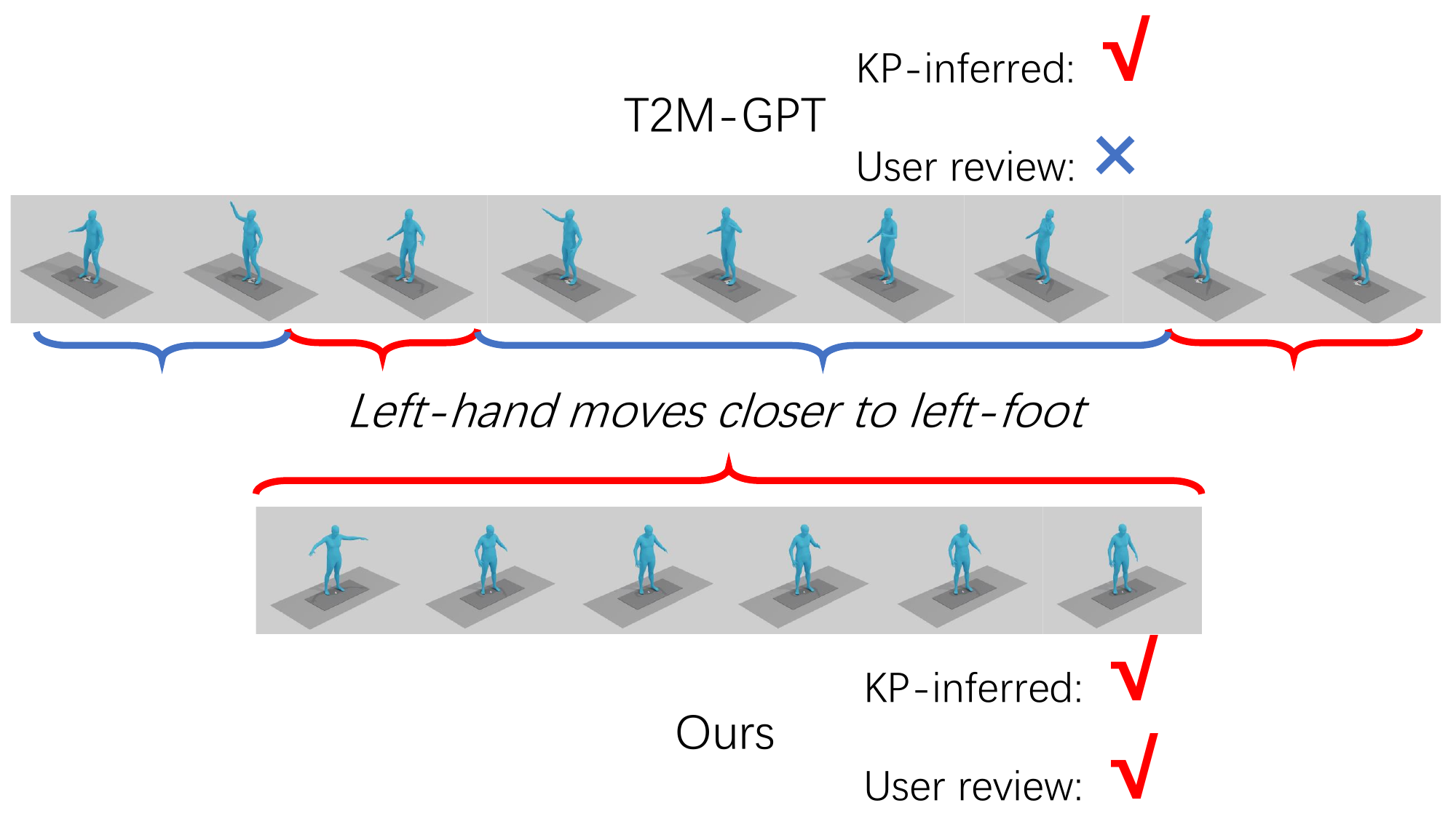}
    \caption{For KPG, we generate more concise motion than T2M-GPT~\cite{t2mgpt}.}
    \label{fig:complex}
\end{figure}

Furthermore, we provide detailed consistency statistics between KP-inferred Accuracy and user-reviewed semantic consistency in Tab.~\ref{tab:kpg-consistency}.
Samples generated from T2M-GPT and our method are included.
KP and users provide similar reviews for over 80\% of the samples, showing good consistency.
Concerning user reviews, KP-inferred Accuracy has a higher false positive rate (0.12 / 0.52 = 0.2308) than a false negative rate (0.04 / 0.48 = 0.0833). 
We find there are two typical false positive scenarios.
First, the generated motion results in rather small indicators, close to the 1e-4 threshold.
KP captures this, however, it is hard for humans to notice such subtle movements.
Second, as shown in Fig.~\ref{fig:complex}, the generated motions sometimes tend to be redundant compared to the given prompts.
Users might be distracted, overlooking the targeted semantics.
We find this happens more for T2M-GPT generated samples (in Fig.~\ref{fig:complex}, extra right-hand waving motion), while our method manages to provide more concise responses.

For the first scenario, we think an adaptive threshold w.r.t. the overall motion intensity would be helpful, since to human perception, the relative amplitude is usually more important than the absolute amplitude. 
Also, extending KP to amplitude might help.
The second scenario urges us to rethink the current text-to-motion task setting. 
For a ``matched'' motion-text pair, should the text semantics be a subset of motion semantics, or strictly match?
Also, is it expected to increase diversity by introducing redundant motions?
We identify these questions as interesting points of attack and leave them for future exploration.

\subsection{Failure Analysis on KPG}

With the attached video \texttt{1286.mp4}, we further demonstrate visualization results on KPG. 
An interesting finding is that different methods show different failure patterns.
Limited motion amplitude is usually observed for MDM~\cite{hmdm} and MLD~\cite{mld}.
Also, MLD~\cite{mld} could misunderstand commands on certain body parts.
T2M-GPT~\cite{t2mgpt} tends to generate over-active motion sequences with redundancy, which could explain its low accuracy for repetitive prompts.
ReMoDiffuse~\cite{ReMoDiffuse} produces jerky motion.
MoMask~\cite{guo2023momask} could surprisingly mystify left with right.
Our model also shows redundancy.
Moreover, most models tend to execute the prompts indirectly, which could indicate the potential over-fitting of motion style.
KPG prompts are simple body-joint relations like exercising instructions; however, they are not usually explicitly described in general datasets. 
Thus, the models tend to produce everyday activity motion, which contains the required relations, instead of directly fulfilling the requirements.
This reveals that current models could be sub-optimal in real understanding of the human body structure. 

\subsection{More Visualizations}
More visualizations are included in the video \texttt{1286.mp4}.

\section{Extensive Discussion}

\textbf{Relation with phase-based methods.} Some previous efforts~\cite{holden2017phase,zhang2018mode,starke2020local} adopted phase-based motion representation, which is similar to Kinematic Phrase in movement representation. However, the term “Phrase” emphasizes the seamless conversion between our phrases and semantic descriptions, which is not explored in previous efforts.

\textbf{Representing complex motions with KP.}
Currently, the KP-based complex motion semantics representation could be conducted demonstration-based. 
That is, given a motion demonstrating Tai-Chi, we could convert it into KPs for a \textit{basic} KP representation of Tai-Chi. 
Then minor modifications could be made to the KP representation for diversity to produce diverse Tai-Chi motions. 
Further exploration of KP-based semantics representation, \textit{e.g.}, the introduction of LLMs might be promising given the symbolic nature of KP. We believe future works on this would be promising.

\textbf{Further exploration on KP-based evaluation.}
Thanks for your constructive comment. A current limitation of KP-aided evaluation is the trade-off between reliability and generality. Initially, we considered comparing the KP similarity of the generated and GT motions for general prompts. However, as the GT might not fully cover expected semantics, this design sacrifices reliability, which is a common issue of previous metrics. Therefore, we limit the current KPG to atomic/two-gram prompts to guarantee reliability and obtain helpful insights. Enhancing KPG with more generality would be promising in future works. Also, KP distribution analysis would be a helpful interpretative analysis tool. 

\textbf{FineMoGen Comparison. } We evaluate it on KPG, with a 46.79\% Acc for simultaneous prompts (ours 43.08\%) and 36.52\% overall Acc (ours 57.86\%). FineMoGen is trained with LLM-extended descriptions similar to the simultaneous prompts. However, it is also biased toward them, resulting in a degenerated overall performance. More details will be updated in the revision.

\end{document}